\documentclass[final]{alt2025}

\title[A PAC-Bayesian Link Between Generalisation and Flat Minima]{A PAC-Bayesian Link Between Generalisation and Flat Minima}
\usepackage{times}

\usepackage{enumitem}
\usepackage{times}
\usepackage{comment}
\usepackage{cleveref}
\usepackage{thmtools}

\crefname{assumption}{Assumption}{Assumptions}
\newtheorem{assumption}[theorem]{Assumption}

\crefname{proposition}{Proposition}{Propositions}
\newtheorem{proposition}[theorem]{Proposition}

\crefname{corollary}{Corollary}{Corollaries}
\newtheorem{corollary}[theorem]{Corollary}

\crefname{definition}{Definition}{Definitions}
\newtheorem{definition}[theorem]{Definition}

\crefname{lemma}{Lemma}{Lemmas}
\newtheorem{lemma}[theorem]{Lemma}

\usepackage{algorithm}
\usepackage{algpseudocode}
\usepackage{nicefrac}

\newcommand{\Ebb}{\ensuremath{\mathbb{E}}}
\newcommand{\Nbb}{\ensuremath{\mathbb{N}}}
\newcommand{\Rbb}{\ensuremath{\mathbb{R}}}
\newcommand{\Pbb}{\ensuremath{\mathbb{P}}}

\newcommand{\wbf}{\mathbf{w}}
\newcommand{\zbf}{\ensuremath{\mathbf{z}}}
\newcommand{\zerobf}{\ensuremath{{\mathbf 0}}}

\newcommand{\Ccal}{\ensuremath{\mathcal{C}}}
\newcommand{\Dcal}{\ensuremath{\mathcal{D}}}
\newcommand{\Bcal}{\ensuremath{\mathcal{B}}}
\newcommand{\Hcal}{\ensuremath{\mathcal{H}}}
\newcommand{\Mcal}{\ensuremath{\mathcal{M}}}
\newcommand{\Ncal}{\ensuremath{\mathcal{N}}}
\newcommand{\Ocal}{\ensuremath{\mathcal{O}}}
\newcommand{\Pcal}{\ensuremath{\mathcal{P}}}

\newcommand{\Scal}{\ensuremath{\mathcal{S}}}
\newcommand{\Tcal}{\ensuremath{\mathcal{T}}}
\newcommand{\Zcal}{\ensuremath{\mathcal{Z}}}

\newcommand{\Drm}{\mathrm{D}}
\newcommand{\Hrm}{\mathrm{H}}
\newcommand{\Lrm}{\mathrm{L}}
\newcommand{\Prm}{\mathrm{P}}
\newcommand{\Qrm}{\mathrm{Q}}

\newcommand{\ie}{\textit{i.e.}\xspace}
\newcommand{\eg}{\textit{e.g.}\xspace}
\newcommand{\wrt}{\textit{w.r.t.}\xspace}
\newcommand{\iid}{\textit{i.i.d.}\xspace}

\newcommand{\defeq}{:=}

\DeclareMathOperator*{\EE}{\Ebb}
\DeclareMathOperator*{\PP}{\Pbb}
\DeclareMathOperator*{\KL}{\mathrm{KL}}
\DeclareMathOperator*{\Ent}{\mathrm{Ent}}
\DeclareMathOperator*{\Var}{\mathrm{Var}}

\newcommand{\Hess}{\mathrm{Hess}}
\newcommand{\Id}{\mathrm{Id}}
\newcommand{\Lsob}{\texttt{L-Sob}}
\newcommand{\Poinc}{\texttt{Poinc}}
\newcommand{\Risk}{\mathrm{R}}
\newcommand{\trace}{\mathrm{trace}}

\newcommand{\D}{\Dcal}
\renewcommand{\H}{\Hcal}
\renewcommand{\P}{\Prm}
\newcommand{\Q}{\Qrm}
\renewcommand{\S}{\Scal}
\newcommand{\z}{\zbf}
\newcommand{\R}{\Rbb}
\newcommand{\N}{\Nbb}

\usepackage{crossreftools}
\renewenvironment{proof}[1][\proofname]{\par\noindent{\bfseries\upshape #1\ }}{\jmlrQED}

\altauthor{%
 \Name{Maxime Haddouche} \Email{maxime.haddouche@inria.fr}\\
 \addr Inria - Département d’Informatique de l’Ecole Normale Supérieure \\
 \addr PSL Research University\\
 \addr Paris, France
 \AND
 \Name{Paul Viallard} \Email{paul.viallard@inria.fr}\\
 \addr Univ. Rennes, Inria, CNRS IRISA - UMR 6074\\
 \addr Rennes, France
 \AND
 \Name{Umut \c{S}im\c{s}ekli} \Email{umut.simsekli@inria.fr}\\
 \addr Inria - Département d’Informatique de l’Ecole Normale Supérieure\\
 \addr PSL Research University\\
 \addr Paris, France  
 \AND
 \Name{Benjamin Guedj} \Email{benjamin.guedj@inria.fr}\\
 \addr Inria and University College London\\ 
 \addr London, UK%
}

\begin{document}

\maketitle

\begin{abstract}%
Modern machine learning usually involves predictors in the overparameterised setting (number of trained parameters greater than dataset size), and their training yields not only good performance on training data, but also good generalisation capacity. This phenomenon challenges many theoretical results, and remains an open problem. To reach a better understanding, we provide novel generalisation bounds involving gradient terms. To do so, we combine the PAC-Bayes toolbox with Poincaré and Log-Sobolev inequalities, avoiding an explicit dependency on the dimension of the predictor space. Our results highlight the positive influence of \emph{flat minima} (being minima with a neighbourhood nearly minimising the learning problem as well) on generalisation performance, involving directly the benefits of the optimisation phase.  
\end{abstract}


\section{Introduction}

\looseness=-1
Understanding generalisation in modern machine learning problems has been a major challenge in learning theory.
The goal here is to upper-bound the so-called \emph{generalisation error}, which is the gap between the population and empirical risks, $\Risk_\D(h) - \hat{\Risk}_{\S_m}(h)$. Here, $h \in \R^d$ represents the parameters of a predictor, $\Risk_{\D}(h)\defeq \EE_{\z\sim \D}[\ell(h,\z)]$ denotes the population risk, $\D$ is an unknown data distribution, $\ell$ is a loss function, $\hat{\Risk}_{\S_m}(h) \defeq \frac{1}{m}\sum_{i=1}^{m} \ell(h,\z_{i})$ is the empirical risk, and $\S_m\defeq \{\z_1,\dots,\z_m\}$ is a dataset in which each $\z_i$ is independent and identically distributed (\iid) with respect to $\D$.

Dating back to \citet{hochreiter1997flat}, it has been hypothesised that the notion of `flatness' (or sometimes equivalently referred to as `sharpness') is closely linked to the generalisation error: among the minima found by the learning algorithm, the flatter the minimum, the lower the generalisation error.
While the initial concept of flatness was (vaguely) defined through low Kolmogorov complexity, there is no globally accepted definition of flatness.
Therefore, several notions of flatness have been considered, typically based on the second-order derivatives of the empirical risk around the local minimum found by the learning algorithm, such as $\trace(\nabla_{h}^2 \hat{\Risk}_{\S_m}(h))$~\citep[see \eg][]{jastrzkebski2017three,wen2023sharpness}.

While there have been several attempts to link some form of flatness to generalisation in a mathematically rigorous manner \citep{neyshabur2017exploring,Petzka2021relative,yue2023sharpness,andriushchenko2023modern}, mainly with the framework of `sharpness aware minimisation'~\citep{foret2021sharpness}, it has been shown recently that flat minima do not always imply good generalisation.
In fact, there are scenarios such that the flattest minima result in the worst generalisation performance compared to non-flat ones~\citep{wen2023sharpness}.

In this study, we aim to develop novel links between flatness and generalisation error from a PAC-Bayesian perspective~\citep{shawetaylor1997pac,mcallester1999some}; see also \citep{guedj2019primer,hellstrom2023generalization,alquier2021user} for an introduction.
While existing work involving PAC-Bayes and flatness either uses PAC-Bayes bounds as complexity measures \citep{neyshabur2017exploring,jiang2020fantastic,dziugaite2020search,viallard2024leveraging} or derives bounds for specific algorithms \citep{negrea2019info,neu2021info}, our aim is to derive general PAC-Bayes bounds involving flatness through gradient norm.
Denoting by $\Q$, the probability distribution of the algorithm's output $h$, we identify sufficient conditions on $\Q$ such that flatness always implies good generalisation.
More precisely, we make the following contributions:
\begin{itemize}
    \item We show that, when $\Q$ satisfies the Poincaré inequality and a technical condition that we identify, we can obtain a `time-uniform estimation' PAC-Bayes bound that mainly contains two terms:  
    \begin{enumerate}[label=(\roman*)]
        \item The flatness term: either $\EE_{\z\sim\D}\EE_{h\sim \Q}[\|\nabla_h\ell(h,\z)\|^2]$ or $\EE_{h\sim \Q}[ \frac{1}{m}\sum_{i=1}^m \|\nabla_h\ell(h,\z_i)\|^2 ]$.
        The latter is directly linked to the Hessian of the loss $\ell$, due to the connection between the Fisher information and the Hessian of the loss \citep{bickel2015mathematical}.
        For instance, under certain conditions, it can be shown that the trace is linked to gradients as $\trace(\nabla_{h}^2 \hat{\Risk}_{\S_m}(h)) = \frac{2}{m}\sum_{i=1}^m \|\nabla_h \ell(h,\z_i)\|^2$ \citep[Lemma 4.1]{wen2023sharpness}.  
        \item The classical PAC-Bayesian complexity term $\KL(\Q,\P)$, where $\KL$ denotes the Kullback-Leibler divergence and $\P$ is data-independent `prior' distribution. 
    \end{enumerate}
    \item We further analyse the term $\KL(\Q,\P)$.
    We show that when $\Q$ is a Gibbs distribution, \ie $d\Q(h)\propto \exp(- \gamma \hat{\Risk}_{\S_m}(h)) d\P(h)$ for some $\gamma >0$ and $\P$ satisfies a log-Sobolev inequality, the generalisation error can be controlled \emph{solely} by the term: $\gamma^2 c_{LS}(\P)\EE_{h\sim \Q}[ \|\nabla_h \hat{\Risk}_{\S_m}(h) \|^2 ]$, where $c_{LS}(\P)$ denotes the log-Sobolev constant of the prior $\P$. 
    \item We go beyond KL divergence to link flat minima to deterministic predictors (\ie when $\Q$ is a Dirac distribution) through a novel Wasserstein-based generalisation bound for gradient Lipschitz loss functions. 
\end{itemize}
We provide a numerical assessment of the technical condition underlying our main result, suggesting that it is suitable for neural networks on classification tasks, confirming the relevance of our bounds to better understand the generalisation ability of such models.
Our results further shed light on the impact of the flatness of minima on generalisation error: when the learning algorithm ensures a sufficiently regular distribution over the parameters, the generalisation error can be directly controlled by the flatness of the region found by the algorithm.

\section{Preliminaries}

\textbf{Framework.}
We consider a predictor set $\H\subseteq \R^d$ equipped with a norm $\|\cdot\|$, a data space $\Zcal$ and the space of distributions $\Mcal(\H)$ over $\H$.
We also consider a loss function $\ell : \H\times \Zcal \rightarrow \R$. 
We assume that we have an \iid dataset $\S=(\z_i)_{i\geq 1}\in\Zcal^\N$  with associated unknown data distribution $\D$. 
For each $m\geq 1$, we define $\S_m\defeq \{\z_1,\dots,\z_m\}$.
In PAC-Bayes learning, we construct a data-driven posterior distribution $\Q\in\Mcal(\H)$ with respect to a prior distribution $P$. 
To assess the generalisation ability of a predictor $h\in\H$, we define the \emph{population risk} to be $\Risk_{\D} (h) \defeq \EE_{\z\sim \D}[\ell(h,\z)]$ and for each $m$, its empirical counterpart as $\hat{\Risk}_{\S_m} (h) \defeq \frac{1}{m}\sum_{i=1}^{m} \ell(h,\z_{i})$. 
We also define the expected and empirical risks for $\Q\in\Mcal(\H)$ as $\Risk_{\D}(\Q)\defeq \EE_{h \sim \Q}[ \Risk_{\D}(h)]$ and $\hat{\Risk}_{\S_m}(\Q)\defeq \EE_{h \sim \Q}[ \hat{\Risk}_{\S_m}(h)]$.
PAC-Bayes bounds usually aim to control the \emph{expected generalisation error} for each dataset size $m$, \ie  $\Risk_{\D}(\Q) - \hat{\Risk}_{\S_m}(\Q)$.
\\

\noindent{}\textbf{Background on Poincaré and log-Sobolev inequalities.} In this work, we exploit Poincaré and log-Sobolev inequalities in the PAC-Bayes framework.
We first recall their definitions: for a fixed distribution $\Q\in\Mcal(\H)$, we define the \emph{Sobolev space of order $1$} on $\R^d$ as follows:
\begin{align*}
\Hrm^{1}(\Q) \defeq \left\{ f\in \Lrm^2(\Q)\cap \Drm_1(\R^d)\mid \|\nabla f\|\in \Lrm^2(\Q) \right\},
\end{align*}
where $\Drm_1(\R^d)$ denotes the set of derivable functions $f : \R^d \to \R$ and $\Lrm^2(\Q)$ is the space of square-integrable function. 
In other words, $\Hrm^{1}(\Q)$ is the set of functions that are square-integrable, with their gradient’s norm also being square-integrable.

\begin{definition}[Poincaré inequality]\label{def:poincare}
A distribution $\Q$ satisfies a \emph{Poincaré inequality} with constant $c_{P}(\Q)$ if for all function $f\in \Hrm^{1}(\Q)$ we have 
\begin{align*}
 \Var_{h\sim\Q}(f(h)) \le c_{P}(\Q) \EE_{h\sim \Q}\left[ \|\nabla_{h} f (h)\|^2 \right],
\end{align*}
where $\Var_{h\sim\Q}(f(h)) = \EE_{h\sim\Q} [f(h) - \EE_{h\sim\Q}[f(h)]]^2$ is the \emph{variance} of $f$ with respect to $\Q$.
We then say that $\Q$ is Poincaré with constant $c_{P}(\Q)$, or that $\Q$ is $\Poinc(c_{P})$.
\end{definition}

\begin{definition}[Log-Sobolev inequality]
A distribution $\Q$ satisfies a \emph{log-Sobolev inequality} with constant $c_{LS}(\Q)$ if for all function $f\in \Hrm^{1}(\Q)$ we have 
\begin{align*}
\Ent_{h\sim\Q}(f^2(h)) \defeq \EE_{h\sim\Q}\left[ f^2(h)\log\left( \frac{f^2(h)}{\EE_{h\sim\Q}\left[ f^2(h)\right]}\right) \right] \le c_{LS}(\Q) \EE_{h\sim \Q}\left[ \|\nabla_{h} f (h)\|^2 \right],
\end{align*}
where the term $\Ent_{h\sim\Q}(f^2(h))$ is the \emph{entropy} of $f^2$.
We then say that $\Q$ is log-Sobolev with constant $c_{LS}(\Q)$, or that $\Q$ is $\Lsob(c_{LS})$.
\end{definition}
The class of Gaussian distributions is an important particular case of distributions satisfying both Poincaré and log-Sobolev inequalities; this is the subject of \Cref{prop:gaussian-inequalities}.
\begin{proposition}[\citet{gross1975logarithmic,brascamp1976extensions,beckner1989generalized}]\label{prop:gaussian-inequalities}
\looseness=-1
Given a distribution $\Q= \Ncal(\mu, \Sigma)$, where $\mu$ is the mean and $\Sigma$ is the covariance matrix in $\R^d$.
Then, for any $f \in \Hrm^{1}(\Q)$:
\begin{align*}
\Ent_{h\sim\Q}(f^2(h)) \le 2\EE_{h\sim\Q}\left[ \left\langle \Sigma \nabla_{h} f(h), \nabla_{h} f(h)\right\rangle \right], \quad\text{and}\quad \Var_{h\sim\Q}(f(h)) \le \EE_{h\sim\Q}\left[ \left\langle \Sigma \nabla_{h} f(h), \nabla_h f(h)\right\rangle \right].
\end{align*}
Thus, the distribution $\Q$ is $\Lsob(c_{LS})$ with constant $c_{LS}(\Q)=2\|\Sigma\|_{op}$ and is also $\Poinc(c_{LS})$ with constant $c_{LS}(\Q)=\|\Sigma\|_{op}$, where $\|\cdot\|_{op}$ denotes the operator norm. 
\end{proposition}
In \Cref{prop:gaussian-inequalities}, the first inequality can be derived from the classical log-Sobolev inequality for $\Ncal(\zerobf,\textrm{Id})$ stated in \citet{gross1975logarithmic}, with a change of variable. Similarly, the Poincaré inequality can be obtained through a change of variable from the Poincaré inequality for $\Ncal(\zerobf,\textrm{Id})$ which is a particular case of the Brascamp-Lieb inequality for log-concave probability measures \citep{brascamp1976extensions} and is stated explicitly in \citet[Theorem 1]{beckner1989generalized}.\\
\noindent{}We now focus on specific posterior distributions called \emph{Gibbs posteriors, or Gibbs distributions}.
Given a fixed loss $\ell : \H\times \Zcal \rightarrow \R$, a prior $\P\in\Mcal(\H)$ and a dataset $\S_m$, the Gibbs posterior $\Q^{\gamma}_{\S_m}$ is defined such that $d\Q^{\gamma}_{\S_m}(h)\propto \exp( - \gamma \hat{\Risk}_{\S_m}(h) ) dP(h)$, where $\gamma>0$ is an \emph{inverse temperature}.
Gibbs posteriors are a class of closed-form solutions for relaxation of \citet[Theorem 1.2.6]{catoni2007pac} stated, for instance, in \citet[Theorem 4.1]{alquier2016properties}.
\Cref{prop:gibbs-logsob} shows that when the prior and the loss satisfy a few properties, then the associated Gibbs posterior is $\Lsob(c_{LS})$.

\begin{restatable}{proposition}{propgibbslogsob}\label{prop:gibbs-logsob}
Assume that $\P$ is a probability measure on $\H$ such that $dP(h) \propto \exp(-V(h))$ with $V:\H\to\R$ a smooth function such that $\Hess(V)\succeq \frac{2}{c_{LS}(\P)}\Id$.\footnote{ The notation $A \succeq B$ means that $A-B$ is a semi-definite positive matrix.} 
Assume that $\ell(h, \z) = \ell_1(h, \z) + \ell_2(h, \z)$ with $\ell_1$ convex, twice differentiable and $\ell_2$ bounded. 
Then for any $\gamma>0$, the Gibbs posterior $\Q^{\gamma}_{\S_m}$ is $\Lsob(c_{LS})$ with constant $c_{LS}(\Q^{\gamma}_{\S_m})= c_{LS}(\P)\exp\left( 4\|\ell_2\|_{\infty}\right)$.
\end{restatable}
\noindent{}\Cref{prop:gibbs-logsob} applies, \eg when $\P$ is a Gaussian prior $\P=\Ncal(\mu_\P,\Sigma_\P)$. Notice that in this case $c_{LS}(\P)= 2\|\Sigma_\P\|_{op}$. This property is a straightforward application of \citet[Corollary 2.1]{chafai2004entropies} with \citet[Property 2.6]{guionnet2003lectures} and is stated in \Cref{sec:supp-background} for completeness.
Finally, notice that satisfying a log-Sobolev inequality is stronger than satisfying a Poincaré one. This is stated for instance in \citet[Proposition 2.1]{ledoux2006concentration} and properly recalled in \Cref{sec:supp-background}. 

\section{Reaching a flat minimum allows Poincaré posteriors to generalise well}
\label{sec:poincare-gauss}

In this section, we consider posterior distributions $\Q$ that are $\Poinc(c_\P)$.
This assumption covers the important case of Gaussian measures (\Cref{prop:gaussian-inequalities}) as well as all measures satisfying a log-Sobolev inequality (\Cref{prop:ls-implies-poinc}).

\subsection{Time-uniform estimation PAC-Bayes bounds for heavy-tailed losses}  
\label{sec:fast-rates-gradient-h}

\looseness=-1
We now focus on \emph{time-uniform estimation} PAC-Bayes bounds, \ie bounds such that there exists $\Ccal \subseteq \Mcal(\H)$, and $\alpha>0$ with probability at least $1-\delta$, for all $\Q\in \Ccal$, and $m>0$, there exists $\varepsilon_m>0$ s.t.
\begin{align*}
\Risk_\D(\Q) \le   \frac{\alpha}{m}\left[\KL(\Q,\P) +\log(1/\delta)\right] + \varepsilon_m.
\end{align*}
\looseness=-1
\noindent Here, 'time-uniform' means that the bound holds with probability $1-\delta$ for all $m$, and 'estimation' means that we directly control $\Risk_\D$ instead of the generalisation gap $\Risk_\D -\hat{\Risk}_{\S_m}$. In particular, if $\sup_{m\geq m_0} \varepsilon_m \le \varepsilon$ for some small $\varepsilon>0$, we interpret a time-uniform estimation bound as a \emph{transitory fast rate}, \ie a bound decaying for all $m\geq m_0$ below $2\varepsilon$ at speed $1/m$. Such a property is of interest to understand why deep neural networks rapidly acquire a good generalisation ability.
Another fundamental difference between time-uniform PAC-Bayes bounds (which recently appeared in \citealp{haddouche2023pac,chugg2023unified}) is that they are linked to almost surely convergence while classical PAC-Bayes results are related to in-probability convergence. We elaborate on this fundamental difference in 
\Cref{sec:fundamental-diff-time-unif}.
\\
To obtain time-uniform estimation bounds, we exploit the notion of flat minima, \ie a minimum whose neighbourhood almost minimises the loss, and this property can be attained in an overparameterised setting such as neural networks once the optimisation phase has been performed.
We exploit this flatness property through the gradient norm $\| \nabla_h \ell(h,\z)\|$ of the loss \wrt the predictor $h$ for any $\z$.
To our knowledge, this is the first attempt to do so, as \citet{gat2022importance} focus on gradients with respect to the data $\nabla_{\z}\ell(h,\z)$.

We first state in \Cref{as:relaxed-bounded} a key assumption of our work, which intricates the data distribution $\D$ with the posterior of interest $\Q$.

\begin{assumption}
\label{as:relaxed-bounded}
We then say that $\Q\in\Mcal(\H)$ is \emph{quadratically self-bounded} \wrt the loss function $\ell:\H\times\Zcal\to\R$ and the constant $C>0$ (namely $\texttt{QSB}(\ell,C)$) if
\begin{align*}
\EE_{\z\sim \D}\left[ \left(\EE_{h\sim\Q}[\ell(h,\z)]\right)^2 \right] \le C \Risk_\D(\Q) = C\EE_{\z\sim \D}\left[ \EE_{h\sim\Q}[\ell(h,\z)] \right].
\end{align*}
\end{assumption}
\Cref{as:relaxed-bounded} is a relaxation of boundedness, as if $\ell: \H\times\Zcal\to[0,C]$ then it is $\texttt{QSB}(\ell,C)$.
It is an alternative to the bounded expected variance assumption in anytime-valid PAC-Bayes bounds \citep{haddouche2023pac,chugg2023unified}.
A key issue with their boundedness assumption is that it must hold for all posteriors, including those providing poor generalisation performance.
Our $\texttt{QSB}$ assumption avoids this by intricately linking the properties of the distribution $\D$, the loss $\ell$ and the posterior $\Q$.
Such a design is in line with the conclusions of the recent work of \citet{gastpar2023fantastic}, inviting to derive generalisation bounds valid for specific pairs $(\Q,\D)$ (rather than uniformly valid for all such pairs) to reach sharper results.
Finally, we interpret $C$ as a contraction constant that attenuates, on average, the local expansion (governed by variances of $\Q$ and $\D$) of the loss around the mean of $\Q$.
To illustrate the applicability of \texttt{QSB} condition beyond bounded losses, we give a concrete example of an unbounded loss satisfying it. 
\begin{example}
    Assume that for any $\z\in\Zcal$, the loss $\ell(\cdot, \z)$ is unbounded and $L$-Lipschitz, that we are in the realisable case, \ie there exists $h^*\in\H$ such that $\forall \z\in\Zcal$, $\ell(h^*,\z)=0$, and that $\Q$ is an arbitrary distribution with mean $m_\Q$ and standard deviation $\sigma_\Q$ both bounded by a certain $K$. Then, since we have $\ell(h,\z)= \ell(h,\z)-\ell(h^*,\z)$, and by Lipschitzness and Cauchy–Schwarz's inequality, for any $\z\in\Zcal$, we can deduce that we have $\EE_{h\sim\Q}[\ell(h,\z)]^2 \le \left(\EE_{h\sim\Q}[\sqrt{L \|h-h^*\|\ell(h,\z)}]\right)^2 \le L\EE_{h\sim\Q}[\|h-h^*\|] \EE_{h\sim\Q}[\ell(h,\z)]$.
Finally, note that by Jensen's inequality and the bias-variance decomposition, we have 
$\EE_{h\sim\Q}[\|h-h^*\|] \le \sqrt{\EE_{h\sim\Q}[\|h-h^*\|^2]} = \sqrt{\sigma_{\Q}^2 +\|m_\Q - h^*\|^2}\le \sqrt{K^2 + (K+\|h^*\|)^2}$. 
This ensures that the QSB condition holds in this case with constant $C= L\sqrt{K^2 + (K+\|h^*\|)^2}$. 
\end{example}
We are now able to state the main result of this section.

\begin{theorem}\label{th:poincare-gauss}
\looseness=-1
For any $C{>}0$, for any $\lambda$ such that $\frac{2}{C}{>}\lambda{>}0$, for any data-free prior $\P\in\Mcal(\H)$, for any loss function $\ell: \H\times\Zcal\to \R_{+}$, and for any $\delta\in (0,1]$, we have, with probability at least $1-\delta$ over the sample $\S$, for all $m\in\N^{*}$, for all $\Q$ being $\Poinc(c_\P)$, $\texttt{QSB}(\ell,C)$, and $\ell(\cdot,\z)\in \Hrm^{1}(\Q)$ for all $\z$,
\begin{multline*}
\Risk_{\D}(\Q) \le \frac{1}{1-\frac{\lambda C}{2}} \left( \hat{\Risk}_{\S_m}(\Q) + \frac{\KL(\Q,\P) +\log(1/\delta)}{\lambda m} \right) \\
+ \frac{\lambda}{2-\lambda C} c_{P}(\Q)\EE_{\z\sim\D} \left[ \EE_{h\sim \Q}\left( \|\nabla_h \ell(h,\z)\|^2 \right) \right].
\end{multline*}
\end{theorem}
\looseness=-1
\Cref{th:poincare-gauss} provides a time-uniform estimation bound with $\alpha=\frac{2}{\lambda(2-\lambda C)}$ and with the threshold $\varepsilon(\Q,\S_m)= \frac{1}{2-\lambda C}\left( 2\hat{\Risk}_{\S_m}(\Q) + \lambda c_\P(\Q)\EE_{\z\sim\D}[\EE_{h\sim \Q}(\|\nabla_h  \ell(h,\z)\|^2)]\right)$ for any $m\in\N^{*}$. Achieving a small $\varepsilon_m$ (and thus approaching a fast convergence rate) requires two conditions: $\hat{\Risk}_{\S_m}(\Q)\approx 0$ and expected gradients to vanish.
While the first condition is often satisfied by deep neural networks, the second holds if a flat minimum has been reached through the optimisation process.
Then, setting $\lambda = \nicefrac{1}{C}$ ensures a transitory fast-rate bound of $\nicefrac{1}{m}$ for any $m \in \N^{*}$.
Otherwise, for a fixed $m$, setting $\lambda=\nicefrac{m^{-\alpha}}{C}$ with $\alpha \in \left[0;\nicefrac{1}{2}\right]$ allows adapting the rate with respect to the behaviour of the gradients.
In the case of constant gradients, we recover a convergence rate of $\nicefrac{1}{\sqrt{m}}$, at the cost of the time-uniform property, matching \citet[Theorem 4.1]{alquier2016properties}. 

\looseness=-1    
\paragraph{On the role of flat minima in PAC-Bayes learning.} 
We highlight that the gradient term in \Cref{th:poincare-gauss} is derived with respect to a predictor $h\in\H$ and not $\z\in\Zcal$ which is, to our knowledge, novel in PAC-Bayes.
This is particularly impacting, as $\nabla_h\ell(h,\z)$ is the gradient involved in learning procedures and, when averaged over $\Q$, provides information about the nature of the minima reached and its neighbourhood; we elaborate further in \Cref{sec:comparison-tolstikhin}.
\Cref{th:poincare-gauss} suggests that, to attain good generalisation ability, the mean of $\Q$ must be close to two minima: 
\emph{(i)} on $\hat{\Risk}_{\S_m}(\Q)$ in order to make $\hat{\Risk}_{\S_m}(\Q)$ small, and \emph{(ii)} on $\EE_{\z\sim\D}[\|\nabla_h \ell(h,\z)\|^2]$ to ensure the gradients are small. The variance of $\Q$ must fit the flatness of these minima to reduce the expected terms on the right-hand side of \Cref{th:poincare-gauss}. Finally, the KL term invites, for Gaussian distributions, considering high variances and flat minima to maintain a small value for the bound.

\paragraph{A focus on $C$.}
Taking $\lambda= \nicefrac{1}{C}$ in \Cref{th:poincare-gauss} reduces the influence of the prior distribution $\P$ while amplifying the gradient term. Therefore, a small $C$ is desirable when working with flat minima to mitigate the effects of a poorly chosen prior. Having a small $C$ is reachable in practice: we show in \Cref{sec:expes}, for a classification task on MNIST, that the \texttt{QSB} assumption holds with $C$ strictly smaller than $1$ when considering neural networks.\\

\begin{proof}[Proof of \Cref{th:poincare-gauss}]
\looseness=-1
We start from \citet[Corollary 17]{chugg2023unified} instantiated with a single $\lambda$, an \iid dataset and a prior $\P$. 
With probability at least $1-\delta$, for all $\Q\in\Mcal(\H)$ and $m\in\N^{*}$, we have
\begin{align*}
\Risk_{\D}(\Q) \le  \hat{\Risk}_{\S_m}(\Q) + \frac{\KL(\Q,\P) +\log(1/\delta)}{\lambda m} 
+ \frac{\lambda }{2}\left(   \EE_{h\sim \Q}\left[\EE_{\z\sim \D}[\ell (h,\z)^2]  \right]  \right),
\end{align*} 
where $\z\sim \D$ is independent from $\S$. 
We study the term $\EE_{h\sim \Q}[\EE_{\z\sim \D}[\ell (h,\z)^2]]$ on the right-hand side. We first apply Fubini's theorem to obtain 
\begin{align*}
\EE_{h\sim \Q}\left[\EE_{\z\sim \D}[\ell (h,\z)^2] \right] & = \EE_{\z\sim \D}\left[ \EE_{h\sim \Q}[\ell (h,\z)^2] \right] = \EE_{\z\sim\D} \left[ \Var_{h\sim \Q}\left( \ell(h,\z) \right) + \left( \EE_{h\sim\Q}[\ell(h,\z)] \right)^2 \right].
\end{align*}
As for any $\z\in\Zcal$, we have $\ell(\cdot,\z)\in \Hrm^{1}(\Q)$, we apply Poincaré inequality to obtain
\begin{align*}
\EE_{h\sim \Q}\left[\EE_{\z\sim \D}[\ell (h,\z)^2] \right] \le  \EE_{\z\sim\D} \left[ c_{P}(\Q)\EE_{h\sim \Q}\left( \|\nabla_h \ell(h,\z)\|^2 \right) + \left( \EE_{h\sim\Q}[\ell(h,\z)] \right)^2 \right]  .
\end{align*}
Using that $\Q$ is $\texttt{QSB}(\ell,C)$ and re-organising the terms gives
\begin{multline*}
\Risk_{\D}(\Q) \le \frac{1}{1-\frac{\lambda C}{2}} \left( \hat{\Risk}_{\S_m}(\Q) + \frac{\KL(\Q,\P) +\log(1/\delta)}{\lambda m} \right) \\
+ \frac{\lambda}{2-\lambda C} c_{P}(\Q)\EE_{\z\sim\D} \left[ \EE_{h\sim \Q}\left( \|\nabla_h \ell(h,\z)\|^2 \right) \right]. 
\end{multline*}
\end{proof}
\noindent When the \texttt{QSB} assumption is not verified, it is still possible to exploit the benefit of flat minima in PAC-Bayes at the cost of an upper bound on $\Risk_\D(\Q)$ and a supplementary Poincaré assumption on the data distribution $\D$.  

\begin{restatable}{corollary}{corpoincarepacb}\label{cor:poincare-pacb}
For any $C{>}0$, for any $\lambda$ such that $\frac{2}{C}{>}\lambda{>}0$, for any data-free prior $\P\in\Mcal(\H)$, for any loss function $\ell: \H\times\Zcal\to \R_{+}$ such that $\ell(h,\cdot)$ is $\Ccal^1$ almost everywhere on $\Zcal$, any $\delta\in (0,1]$, if the data distribution $\D$ is $\Poinc(c_\P)$, with probability at least $1-\delta$ over the sample $\S$, for any $m\in\N^{*}$, any posterior $\Q$ being $\Poinc(c_\P)$ with $\Risk_\D(\Q)\le C$ and such that for any $\z\in\Zcal$, $\ell(\cdot,\z)\in \Hrm^{1}(\Q)$:
\begin{multline*}
\Risk_{\D}(\Q) \le \frac{1}{1-\frac{\lambda C}{2}}\left( \hat{\Risk}_{\S_m}(\Q) + \frac{\KL(\Q,\P) +\log(1/\delta)}{\lambda m}\right) \\
+ \frac{\lambda}{2-\lambda C}\left( c_{P}(\Q)\EE_{\z\sim\D} \left[ \EE_{h\sim \Q}\left( \|\nabla_h \ell(h,\z)\|^2 \right) \right]  + c_{P}(\D)\EE_{\z\sim\D} \left( \left\|\EE_{h\sim\Q}[\nabla_\z \ell(h,\z)] \right\|^2\right)  \right).
\end{multline*}
\end{restatable}   
\noindent{}The proof is deferred to \Cref{sec:proof-poincare-pacb}.
\Cref{cor:poincare-pacb} states that if $\Q$ reaches a flat minimum (meaning $\|\nabla_h\ell(h,\z)\|$ is small) and this minimum is robust to the training dataset (meaning $\|\nabla_\z\ell(h,\z)\|$ is small), then a time-uniform estimation bound is attainable with small $\varepsilon_m$, requiring only an upper bound on $\Risk_\D(\Q)$. The assumption of small $\|\nabla_\z\ell(h,\z)\|$ can be attained with algorithms such as Sharpness-Aware Minimisation \citep{foret2021sharpness}. Another lead would be to focus on specific predictors where this gradient is directly controlled, as in Lipschitz neural networks training or in adversarial robustness training \citep{madry2018deep,li2019prevent}.
\\
\Cref{cor:poincare-pacb} holds when  $\D$ is $\Poinc(c_\P)$, encompassing the case of Gaussian mixtures \citep{schlichting2019poincare}, which can approximate any smooth density \citep[as recalled in][]{gat2022importance}.
However, the Poincaré constant of a general mixture is unknown, and the upper bound of \citet{schlichting2019poincare} scales with the number of components, involving potentially high $\chi^2$ divergences.\\

\looseness=-1
\noindent\textbf{Comparison with \citet{gat2022importance}}. 
We compare \Cref{cor:poincare-pacb} with \citet[Theorems 3.5 and 3.6]{gat2022importance}.
First, our result holds under the assumption that the distribution $\D$ follows a Poincaré inequality, which is strictly less restrictive than assuming a log-Sobolev inequality (\Cref{prop:ls-implies-poinc}).
Second, they assume a bounded loss and focus solely on classification tasks satisfying a technical assumption (see their Lemma 3.3) while ours holds for any learning problem at the sole assumption of a bounded $\Risk_\D(\Q)$, allowing the loss $\ell$ to be unbounded (and non-negative). 
To conclude their proof, \citet{gat2022importance} use a uniform bound on $\EE_{\z\sim\D}[\|\nabla_\z \ell(h,\z)\|]$ in their Theorem 3.5 to have a tractable bound, diminishing the benefits of gradient norm.
While they address this limitation in \citet[Theorem 3.6]{gat2022importance}, the explicit influence of the gradient norm appears within an exponential moment on the losses (attenuated by a logarithm), averaged \wrt the data-free prior $\P$.
Thus, the associated gradients have no apparent reason to be small, and their result cannot be linked to flat minima.
In contrast, \Cref{cor:poincare-pacb} involves expected gradients \wrt the posterior $\Q$, which may reach flat minima. 

\subsection{Towards fully empirical bound for gradient-Lipschitz functions}

In this section,  we assume that the gradient $\nabla_h\ell(h,\z)$ is $G$-Lipschitz for any $\z\in\Zcal$, which is a classical assumption in optimisation, especially when considering non-convex objectives for SGD \citep{ghadimi2013stochastic,panageas2017gradient,garrigos2023handbook}.
A large portion of high-probability PAC-Bayes bounds are fully empirical, meaning that the right-hand side of the bounds can be computed.
This has numerous advantages, including in-training numerical evaluation of the bound and the development of novel PAC-Bayesian algorithms that minimises such empirical bounds; see \citep{dziugaite2017computing,perezortiz2021progress,viallard2023learning} among others.
However, \Cref{th:poincare-gauss} and \Cref{cor:poincare-pacb} are not fully empirical, as they involve terms such as $\EE_{\z\sim\D}[\EE_{h\sim\Q}[\|\nabla_h \ell(h,\z)\|^2]]$ and $\EE_{\z\sim\D}[ \|\EE_{h\sim\Q}[\nabla_\z \ell(h,\z)]\|^2]$, which involves an expectation over $\z\sim\D$ and are thus not computable in practice. As a result, they lack the desirable properties of fully empirical bounds; we address this issue in \Cref{th:poincare-grad-lpz}.
\begin{restatable}{theorem}{thpoincaregradlpz}
  \label{th:poincare-grad-lpz}
  \looseness=-1
    For any $C_1,C_2,c>0$, for any data-free prior $\P\in\Mcal(\H)$, for any loss function $\ell: \H\times\Zcal\to \R_{+}$ being $\Ccal^2$ and for any $\delta\in (0,1]$, we have, with probability at least $1-\delta$ over the sample $\S$, for all $m\in\N^{*}$, for all $\Q$ being $\Poinc(c_\P){=}c$, $\texttt{QSB}(\ell,C_1)$, $\texttt{QSB}\left(\|\nabla_h\ell\|^2,C_2\right)$, and $\ell(\cdot,\z)\in\Hrm^{1}(\Q)$, and $\|\nabla_h \ell(\cdot,\z)\|^2 \in \Hrm^{1}(\Q)$ for all $\z$, 
    \begin{multline*}
      \Risk_{\D}(\Q) \le  2 \hat{\Risk}_{\S_m}(\Q) + \frac{2c}{C_1} \EE_{h\sim \Q}\left[ \frac{1}{m}\sum_{i=1}^m \|\nabla_h\ell(h,\z_i)\|^2 \right] \\
       + 2\left( C_1 + c\frac{4cG^2 + C_2}{C_1} \right)\frac{\KL(\Q,\P) +\log(2/\delta)}{m}.
    \end{multline*}
\end{restatable}
\looseness=-1
\noindent{}The proof is deferred to \Cref{sec:proof-poincare-grad}.
We showed that to attain fast rates, the \texttt{QSB} assumption must hold for both the loss and its gradient.
We are then able to derive an empirical generalisation bound, involving both empirical loss and gradients.

As \Cref{th:poincare-grad-lpz} is fully empirical, it can be transformed into a generalisation metric, \ie an empirical function of the predictor whose increase or decrease is correlated to the increase or decrease of the generalisation ability of the predictor. In the case of PAC-Bayes, the generalisation metric comes from the generalisation bound.
Such an idea has been exploited recently \citep{neyshabur2017exploring,jiang2020fantastic,dziugaite2020search,viallard2024leveraging} to show that flatness of the empirical risk was correlated to generalisation.
In particular, from $\hat{\Risk}_\S(\Q)$, \citet{neyshabur2017exploring} derived a notion of \emph{sharpness}, stated in \Cref{eq:flat-minima-neyshabur}, which gives information about the flatness of the reached minima for any $\Q=\Ncal(\mu_\Q,\sigma^2 \Id)$.
This notion is defined by 
\begin{align}
  \label{eq:flat-minima-neyshabur}
  \EE_{\nu\sim \Ncal(\zerobf,\sigma^2 \Id)} \left[ \hat{\Risk}_{\S_m}(\mu_\Q + \nu) -  \hat{\Risk}_{\S_m}(\mu_\Q) \right].
\end{align}
Sharpness is then the averaged risk of a predictor drawn under $\Ncal(\mu_\Q,\sigma^2 \Id)$ \wrt its mean $\Q$.
\Cref{th:poincare-grad-lpz} enhance this notion of sharpness with empirical gradients when $\Q$ is $\texttt{QSB}(\ell,C_1)$: 
\begin{equation}
  \label{eq:flat-minima-us}
  \textrm{Sharp}_{\frac{\sigma^2}{C_1}}(\Q)\defeq \EE_{\nu\sim \Ncal(\zerobf,\sigma^2 \Id)} \!\left[ 2\!\left[\hat{\Risk}_{\S_m}(\mu_\Q{+}\nu){-}\hat{\Risk}_{\S_m}(\mu_\Q)\right]{+}\frac{\sigma^2}{C_1}\!\left[\hat{\textrm{G}}_{\S_m}(\mu_\Q{+}\nu){-}\hat{\textrm{G}}_{\S_m}(\mu_\Q)\right] \right]\!,
\end{equation}
where $\hat{\textrm{G}}_{\S_m}(h) = \frac{1}{m}\sum_{i=1}^m\|\nabla_h \ell(h, \z_i)\|^2$.
\looseness=-1
This gradient term can be seen as the norm of an empirical Fisher information, linked to the second-order moment derivative.
Thus, \Cref{eq:flat-minima-us} involves a notion of flatness on both the loss and its gradient, unlike \Cref{eq:flat-minima-neyshabur}.
For the sake of clarity, we specialise \Cref{th:poincare-grad-lpz} in \Cref{cor:flatness-grad-lpz} to Gaussian distributions, introducing this notion of sharpness.
\begin{corollary}\label{cor:flatness-grad-lpz}
For any $C_1,C_2>0$, for any data-free prior $\P=\Ncal(\mu_\P,\sigma^2 \Id)$ with fixed variance $\sigma^2>0$, for any loss function $\ell: \H\times\Zcal\to \R_{+}$ being $\Ccal^2$ for and any $\delta\in (0,1]$, we have, with probability at least $1-\delta$ over the sample $\S$, for all $m\in\N^{*}$, for all $\Q= \Ncal(\mu_\Q,\sigma^2 \Id)$ being $\texttt{QSB}(\ell,C_1)$, $\texttt{QSB}\left(\|\nabla_h \ell\|^2,C_2\right)$ and $\ell(\cdot,\z)\in\Hrm^{1}(\Q)$, and $\|\nabla_h \ell(\cdot,\z)\|^2 \in \Hrm^{1}(\Q)$ for all $\z$, 
\begin{align*}
\Risk_{\D}(\Q) \le 2 \hat{\Risk}_{\S_m}(\mu_\Q) + \hat{\textrm{G}}_{\S_m}(\mu_\Q) + \textrm{Sharp}_{\frac{\sigma^2}{C_1}}(\Q) + \Ocal\left(\frac{\KL(\Q,\P) +\log(2/\delta)}{ m}\right).
\end{align*}
\end{corollary}

\section{Generalisation ability of Gibbs distributions with a log-Sobolev prior}
\label{sec:gibbs}

A limitation of the results in \Cref{sec:poincare-gauss} is that the KL divergence term remains generally uncontrolled, as its formulation depends on the nature of $\P$ and $\Q$.
While a closed form exists for Gaussian distributions, a natural question is whether it is possible to explicitly control the KL term for another class of distributions. 
Following the approach of \citet{catoni2007pac}, we focus in this section on Gibbs posteriors, which naturally arise in PAC-Bayes through the use of tools from statistical physics. We show that log-Sobolev inequalities allow us to control the KL divergence of such distributions with respect to their priors.

\paragraph{Controlling the KL divergence when $\Q$ is a Gibbs posterior.}
\Cref{lemma:kl-bound} exploits the fact that the KL divergence can be expressed as an entropy with respect to the prior distribution $\P$. It shows that the KL divergence of the Gibbs posterior $\Q^{\gamma}_{\S_m}$ is upper-bounded by gradient terms, provided that the prior $\P$ satisfies a log-Sobolev inequality. 
\begin{restatable}{lemma}{lemmaklbound}
\label{lemma:kl-bound}
\looseness=-1
For any $\gamma>0$, for any $m\in\N^{*}$, for any data-free prior $\P\in\Mcal(\H)$ being $\Lsob(c_{LS})$, for any loss function $\ell: \H\times\Zcal\to \R_{+}$ such that $\ell(\cdot,\z) \in \Hrm^{1}(\P)$ for any $\z$, we have
\begin{align*}
\KL\left( \Q^{\gamma}_{\S_m},\P\right) \le \frac{\gamma^2 c_{LS}(\P)}{4} \EE_{h\sim \Q^{\gamma}_{\S_m}}\left[ \|\nabla_h \hat{\Risk}_{\S_m}(h) \|^2 \right].
\end{align*}
\end{restatable}
\noindent{}The proof is deferred to \Cref{sec:proof-kl-bound}. The key message of this lemma is that for Gibbs posteriors, the expansion of the KL divergence is controlled by an expected empirical gradient term. 
Note in this case that, while the KL divergence has an explicit formulation, it requires calculating the exponential moment $\EE_{h\sim \P}[\exp(-\gamma \hat{\Risk}_{\S_m}(h))]$ which is costly in practice. In contrast, we only need to estimate a second-order moment over $\Q^{\gamma}_{\S_m}$.

\paragraph{Generalisation ability of Gibbs posteriors.}
When Gibbs posteriors are involved, the KL divergence can be controlled by a gradient term. An ideal approach, as in \Cref{sec:poincare-gauss}, would be to involve a Poincaré inequality. However, Gibbs posteriors do not necessarily satisfy a Poincaré inequality as in \Cref{sec:poincare-gauss}, so we need to make supplementary assumptions about the loss.

\begin{restatable}{theorem}{thgibbspacb}\label{th:gibbs-pacb}
We first have $C>0$, $\gamma>0$, and a data-free prior $\P\in\Mcal(\H)$ being $\Lsob(c_{LS})$.\\
\emph{(i)} For any loss function $\ell: \H\times\Zcal\to [0,1]$ and for any $\delta\in (0,1]$, with probability at least $1-\delta$ over the sample $\S$, for any $m\in\N^{*}$, we have
\begin{align*}
    \Risk_{\D}(\Q^{\gamma}_{\S_m}) & \le 2 \left( \hat{\Risk}_{\S_m}(\Q^{\gamma}_{\S_m}) + \frac{\gamma^2 c_{LS}(\P)}{4m} \EE_{h\sim \Q^{\gamma}_{\S_m}}\left[ \|\nabla_h \hat{\Risk}_{\S_m}(h) \|^2 \right] + \frac{\log(1/\delta)}{m} \right).
\end{align*}
\emph{(ii)} For any loss function $\ell$ and prior $\P$ satisfying the conditions of \Cref{prop:gibbs-logsob}. Then, for any $\frac{2}{C}>\lambda>0$, with probability at least $1-\delta$ over the sample $\S$, for all $m\in\N^{*}$, assuming $\ell(\cdot,\z)\in \Hrm^{1}(\Q^{\gamma}_{\S_m})$ and that $\Q^{\gamma}_{\S_m}$ is $\texttt{QSB}(C)$, we have
\begin{multline*}
    \Risk_{\D}(\Q^{\gamma}_{\S_m})  \le \frac{1}{1-\frac{\lambda C}{2}} \left( \hat{\Risk}_{\S_m}(\Q^{\gamma}_{\S_m}) + \frac{\gamma^2 c_{LS}(\P)}{4\lambda m} \EE_{h\sim \Q^{\gamma}_{\S_m}}\left[ \|\nabla_h \hat{\Risk}_{\S_m}(h) \|^2 \right] + \frac{\log(1/\delta)}{\lambda m} \right) \\ 
    + \frac{\lambda e^{4\|\ell_2\|_{\infty}}c_{LS}(\P) }{4-2\lambda C} \EE_{\z\sim\D} \left[ \EE_{h\sim \Q^{\gamma}_{\S_m}}\left( \|\nabla_h \ell(h,\z)\|^2 \right) \right] .
\end{multline*} 
\end{restatable}
\looseness=-1
\noindent{}The proof is deferred to \Cref{sec:proof-gibbs-pacb}.
Note that we could have also derived a result analogous to \Cref{cor:poincare-pacb} at the cost of an additional Poincaré assumption on $\D$.
The influence of the inverse temperature $\gamma$ is quadratic: this is the price to pay to fit the dataset and reduce the influence of the prior.
This dependency is attenuated by a gradient term, which is small if a flat minimum on the $\hat{\Risk}_{\S_m}(h)$ has been reached.
This suggests that in the case of Gibbs posteriors with a log-Sobolev prior, reaching a flat minimum on $\hat{\Risk}_{\S_m}(h)$ controls not only $\hat{\Risk}_{\S_m}(\Q)$, but also the KL divergence and this last property is not reachable when considering Poincaré distributions.
The other gradient term comes from \Cref{sec:poincare-gauss} and requires to be close to a flat minimum on $\Risk_\D(h)$  to attain fast rates. 
\paragraph{Comparison to literature.} To our knowledge, existing bounds for Gibbs posteriors do not involve the gradient norm of the posterior. For instance, \citet[Theorem 4.2]{zhang2006info} involves an exponential moment with respect to the prior distribution, while \citet[Equation 8]{kuzborskij2019distribution} consider flatness through ellipsoids around a minimum, controlling it via an 'effective dimension', which is a function of the eigenvalue of the Hessian of the theoretical risk. Finally, \citet[Equation 6]{rivasplata2020pac} propose a bound for bounded losses, removing the KL divergence when Gibbs posteriors are considered, with a rate of $\Ocal(\nicefrac{1}{\sqrt{m}} + \nicefrac{\gamma}{m})$. Assuming the Gibbs posterior reaches a flat minimum such that  $\gamma\EE_{h\sim \Q^{\gamma}_{\S_m}}[ \|\nabla_h \hat{\Risk}_{\S_m}(h) \|^2 ] \le 1$, then \Cref{th:gibbs-pacb} yields a bound of magnitude $\Ocal(\hat{\Risk}_{\S_m}(\Q^{\gamma}_{\S_m})+ \nicefrac{1}{m}+\nicefrac{\gamma}{m})$, representing a transitory fast rate with threshold $\hat{\Risk}_{\S_m}(\Q^{\gamma}_{\S_m})$. 

\section{On the benefits of the gradient norm in Wasserstein PAC-Bayes learning}

\looseness=-1
In \Cref{sec:poincare-gauss,sec:gibbs}, we provided various generalisation bounds, benefiting from flat minima.
However, our results involve a KL divergence, implying absolute continuity of $\Q$ with respect to $\P$, making them incompatible with deterministic predictors (obtained with Dirac distributions). Then the following question arises: can we benefit from the desirable properties of flat minima in a PAC-Bayes bound valid for deterministic predictors?
To address this issue, a recent line of work has emerged, focusing on integral probability metrics, with particular attention to the $1$-Wasserstein distance \citep{amit2022integral,haddouche2023wasserstein,viallard2023learning,viallard2024tighter}; see \Cref{def:wasserstein}.
The idea behind these works is to replace the change of measure inequality  \citep{csizar1975divergence,donsker1976asymptotic} with the Kantorovich-Rubinstein duality \citep{villani2009optimal}, trading a KL divergence for a Wasserstein distance. This not only yields sound theoretical bounds but also PAC-Bayes algorithms for deterministic predictors and Lipschitz neural networks \citep{viallard2023learning}.
We go further here by obtaining the first PAC-Bayesian bound directly involving a $2$-Wasserstein distance, trading a Lipschitz assumption for a gradient-Lipschitz one, which is well-suited for optimisation. To do so, we derive a novel change of measure inequality tailored for the condition $(\star)$ described below, which is a relaxation of the gradient Lipschitz assumption.
\begin{align*}
f: \H \rightarrow \R \;\text{satisfies}\; (\star) \Leftrightarrow  \exists G>0, \forall(a,a')\in \H^2, \langle \nabla f(\cdot),a-a'\rangle\; \text{is $G\|a-a'\|$-Lipschitz}.
\end{align*}

\begin{restatable}{theorem}{thtwowass}
  \label{th:2-wass}
  For any predictor set $\H$ with finite diameter $D>0$, and for any function $f:\H\rightarrow \R$ satisfying $(\star)$, we have for all distributions $\P\in\Mcal(\H)$ and $\Q\in\Mcal(\H)$ 
    \begin{align*}
  \EE_{h\sim \Q}[f(h)] \le \frac{G}{2}W_2^2(\Q,\P) + \EE_{h\sim \P}[f(h)] +D \EE_{h\sim\Q}[\|\nabla f (h)\|].
  \end{align*}
\end{restatable}
\noindent{}The proof is deferred to \Cref{sec:proof-2-wass}.
\Cref{th:2-wass} shows that, when gradients are Lipschitz, it is possible to obtain a duality formula involving the gradient of the considered function, at the cost of a linear dependency on the diameter $D$ of $\H$.
\Cref{th:2-wass} is also linked to the change of measure inequality \citep{csizar1975divergence,donsker1976asymptotic} when the prior distribution satisfies a log-Sobolev inequality. This link is detailed in \Cref{cor:kl-change}.

\begin{restatable}{corollary}{corklchange}
  \label{cor:kl-change}
For any distribution $\P$ being $\Lsob(c_{LS})$ such that $d\P(h) \propto \exp(-V(h))dh$, with $V$ being $\Ccal^2$, for any $R>0$, for any function $f$ on the centred ball $\Bcal(\zerobf,R)$ of radius $R$ satisfying the $(\star)$ assumption, and for any distribution $\Q\in\Mcal(\H)$, we have:
\begin{align*}
\EE_{h\sim \Q}\left[ f\left(\Pcal_R(h)\right) \right] \le \frac{Gc_{LS}(\P)}{4}\KL(\Q,\P) + \EE_{h\sim \P}\left[ f\left(\Pcal_R(h)\right) \right] + 2R\EE_{h\sim \Q} \left[ \left\|\nabla_{h} f \left(\Pcal_R(h)  \right) \right\| \right],
\end{align*}
where $\Pcal_R$ denotes the Euclidean projection onto $\Bcal(\zerobf,R)$.
\end{restatable}
\noindent{}The proof is deferred to \Cref{sec:proof-kl-change}.
\Cref{cor:kl-change} involves a KL divergence and an Euclidean predictor space $\H= \R^d$. This comes at the cost of approximating $\Q$ and $\P$ by, respectively, $\Pcal_R\#\Q$ and $\Pcal_R\#\P$. Thus, the radius $R$ is now a hyperparameter balancing the tradeoff between the quality of our approximations and the looseness of the bound (if the gradient norm is large). A notable strength is that the smoothness assumption is relaxed to apply only within the centred ball $\Bcal(\zerobf,R)$.\\

\noindent{}From \Cref{th:2-wass}, we now derive a novel generalisation bound allowing deterministic predictors.

\begin{restatable}{theorem}{thwpbgrad}
\label{th:wpb-grad}
Let $\delta\in(0,1)$ and $P\in\Mcal(\H)$ a data-free prior.
Assume $\H$ has a finite diameter $D>0$, for any loss function $\ell: \H\times\Zcal\to \R_{+}$ and any $m \in \N^*$, the generalisation gap $h\mapsto\Risk_\D(h)-\hat{\Risk}_{\S_m}(h)$ satisfies $(\star)$.
Assume that $\EE_{h\sim\P}\EE_{\z\sim\D}[\ell(h,\z)^2] \le \sigma^2$, then the following holds with probability at least $1-\delta$, for any $m>0$ and any $\Q \in \Mcal(\H)$:
\begin{align*}
    \Risk_\D(\Q) \le \hat{\Risk}_{\S_m}(\Q) + \frac{G}{2} W_2^2(\Q,\P) + \sqrt{\frac{2\sigma^2\log\left( \frac{1}{\delta} \right)}{m}} + D \EE_{h\sim \Q}\left( \left\| \nabla_h \Risk_\D(h) - \nabla_h \hat{\Risk}_{\S_m}(h) \right\| \right).
\end{align*}
\end{restatable}
\noindent{}The proof is deferred to \Cref{sec:proof-wpb-grad}.  
\Cref{th:wpb-grad} is not the first generalisation bound to involve a $2$-Wasserstein distance \citep{lugosi2022generalization,lugosi2023online}.
However, these results require infinitely smooth loss functions.
Additionally, the results from \citet{amit2022integral,haddouche2023wasserstein,viallard2023learning}, which use the $1$-Wasserstein, can be directly relaxed on bounds involving the $2$-Wasserstein, while still requiring a Lipschitz loss.
In contrast, our result holds for any nonnegative loss whom the generalisation gap $ h\rightarrow\Risk_\D(h)-\hat{\Risk}_{\S_m}(h)$, satisfies $(\star)$, being a relaxation of gradient-Lipschitz assumption.
\Cref{th:wpb-grad} involves a rate of $\nicefrac{1}{\sqrt{m}}$, as we have to control the generalisation gap over $\P$.
Another restriction of our result, compared to previous ones, is that it holds for $\H$ with a finite diameter. 

\paragraph{Can \Cref{th:wpb-grad} go to zero with large $m$?} In its current form, it is unclear whether \Cref{th:wpb-grad} goes to zero with large $m$, as the Wasserstein distance and the gradient term, have no explicit convergence rate in $m$. Concerning the gradient term, it has been shown that, if we consider $\Q$ to be a Dirac in the output of SGD after $T$ iterations, then according to \citet[Theorem 3]{li2022sgd}, when the intrinsic noise of SGD is subgaussian, the output $\wbf_T\in\R^d$ of SGD satisfies with high probability:
$\|\nabla_h \Risk_{\S_m}(\wbf_T)- \nabla_h \Risk_{\D}(\wbf_T)\|^2 \le \Ocal\left( \frac{d\sqrt{T}\log(T)}{m}\right)$. Concerning the Wasserstein term, if we assume directly that $\ell$ satisfies $(\star)$ with constant $G$, then using a technique inspired from \citet{amit2022integral} yields with probability $1-\delta$ that the generalisation gap satisfies $(\star)$ with constant $G'=\Ocal\left( G\sqrt{\nicefrac{\log(|\H|/\delta)}{m}}\right)$ when $\H$ is finite. We prove this in \Cref{sec:wass-to-zero}.

\section{An empirical study of \Cref{as:relaxed-bounded} for neural networks\protect\footnote{The source code is available at this \href{https://github.com/paulviallard/ALT25-PAC-Bayes-Flat-Minima}{link}.}}
\label{sec:expes}
In this section, we empirically verify whether the \texttt{QSB} assumption holds for neural nets.
This allows us to assess whether \Cref{th:poincare-gauss} helps in understanding the generalisation ability of neural nets.\\

\noindent\textbf{Experimental protocol.} 
We consider classification tasks on two datasets: MNIST~\citep{lecun1998mnist} and FashionMNIST~\citep{xiao2017fashion}.
We have kept the original training set $\S_m$ and the original test set denoted by $\Tcal_n$ (of size $n$).
We consider the convolutional neural network of \citet{springenberg2015striving} adapted for MNIST and FashionMNIST.
The model is composed of $4$ layers containing $10$ channels with a $5{\times}5$-kernel; we set the stride and the padding to $1$, except for the second layer, where it is fixed to $2$. 
Each of these (convolutional) layers is followed by a Leaky ReLU activation function.
Moreover, an average pooling with a $8{\times}8$-kernel is performed before the Softmax activation function.
To initialise the weights of the network, we use \citet{glorot2010understanding} uniform initialiser, while the biases are initialised in $[-\nicefrac{1}{\sqrt{250}}, +\nicefrac{1}{\sqrt{250}}]$ uniformly (except the first layer, the interval is $[-\nicefrac{1}{5}, +\nicefrac{1}{5}]$).
Hence, in this case, $\H$ is the set of neural networks with a fixed architecture, and parametrised with a vector $\wbf$.
The posterior distribution $\Q$ is a Gaussian measure $\Ncal(\wbf, \sigma^2\textrm{Id})$ centred on the parameters $\wbf$ associated with the model; $\sigma$ is set to $10^{-4}$. 
Note that this distribution respects the $\Poinc(c_\P)$ assumption; see \Cref{sec:fast-rates-gradient-h}.
We train the neural network with the (vanilla) stochastic gradient descent algorithm, where the batch size is equal to $512$, and the learning rate is fixed to $10^{-2}$.
We train for at least $10^{4}$ gradient steps and finish the current epoch when this number of iterations is reached.
Our loss $\ell$ is the bounded cross-entropy loss of~\citet[Section D]{dziugaite2017computing}.

\begin{figure}[!ht]
    \centering
    \includegraphics[width=1.0\linewidth]{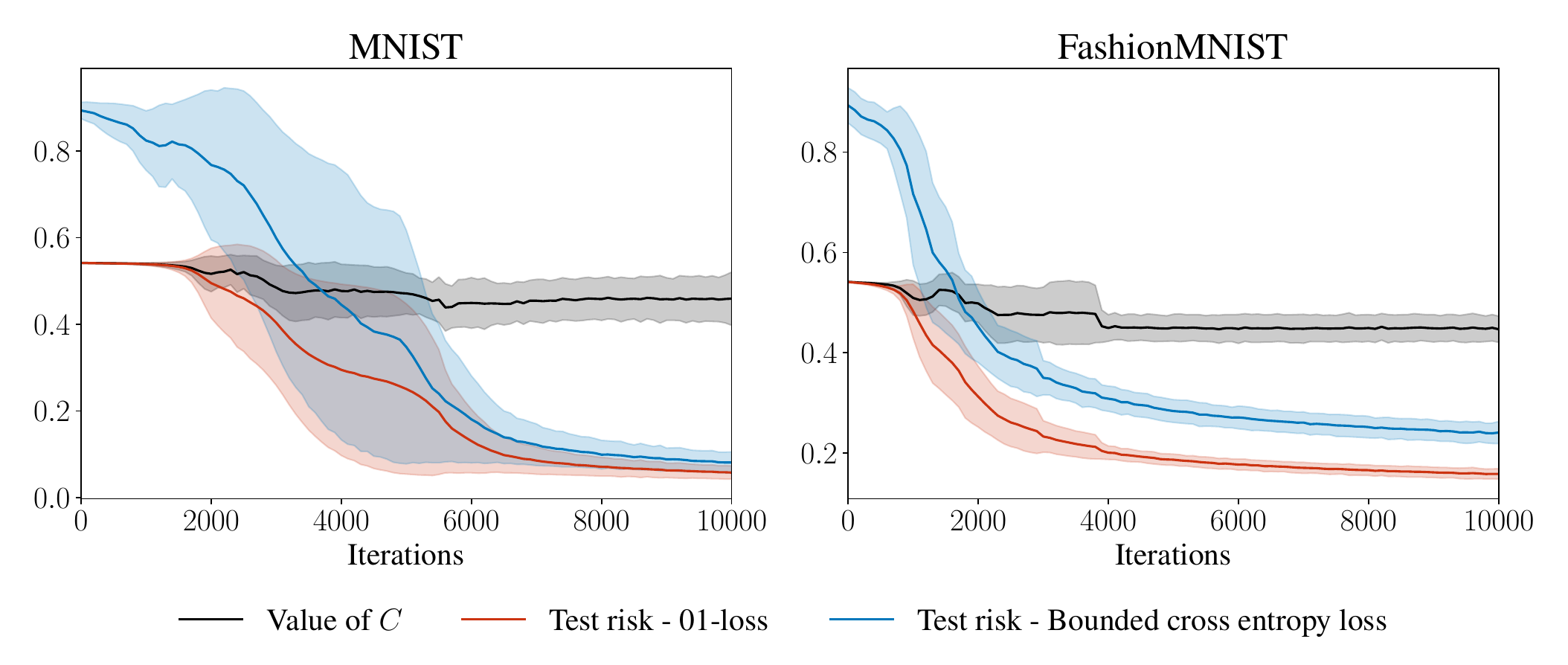}
    \caption{Evolution of the test risks (with the $01$-loss and the bounded cross-entropy loss) and the value of $C$ during the training phase.}
    \label{fig:expe}
\end{figure}

In \Cref{fig:expe}, we report the evolution of three quantities: \emph{(i)} the estimated value of $C$, \emph{(ii)} the test risk $\hat{\Risk}_{\Tcal_n}(\Q)$ and \emph{(iii)} the test risk with the 01-loss.
More precisely, the risks and $C$ are estimated by sampling 10 hypotheses from $\Q$ and by computing the values on a mini-batch of $\Tcal_n$ (with $512$ examples) at each iteration.
Then, \Cref{fig:expe} represents averaged values on 5 runs, each point of the curve representing the average on 100 iterations of the training process (for $10^{4}$ iterations, we only plot $10^{2}$ averaged points for clarity).

\paragraph{Empirical findings.}
\Cref{fig:expe} illustrates that, when neural networks are involved for two classification tasks, $\Q$ evolves during the optimisation process while maintaining the \texttt{QSB} property with constant $C<1$.
For both MNIST and FashionMNIST, the constant $C$ decreases from approximately 0.55 to 0.45. We deduce that having a data-free $\P$ (0 iteration) being \texttt{QSB} with $C<1$ suggests that the architecture of our neural network also has an influence on the $\texttt{QSB}$ assumption. As specified in \Cref{sec:poincare-gauss}, having $C<1$ attenuates the impact of the KL term, thus $\P$. This is desirable as it allows the optimiser to deeply explore the predictor space when $\P$ yields poor performances. We also note that the generalisation ability of $\Q$ on the training loss nearly matches the performance on the 0-1 loss for MNIST but is deteriorated for FashionMNIST, this invites to study more deeply the design of such surrogates in future work. 

Finally, the take-home message of this study is that the \texttt{QSB} assumption is verified for small neural networks on MNIST. Such an empirical confirmation is crucial as it is required for our main result (\Cref{th:poincare-gauss}) and thus confirms that, for neural networks, reaching flat minima during the optimisation phase translates in increased generalisation ability.

\section{Conclusion}

We provide novel time-uniform PAC-Bayes bounds, that can be interpreted as a transitory convergence rate of $\nicefrac{1}{m}$ when a low empirical error is reached and expected gradients vanish. Doing so, we draw sound theoretical links highlighting the impact of flat minima in generalisation.
However, a crucial open question remains: how do optimisation algorithms successfully attain flat minima in the overparameterised setting?
We leave this important question for future work.

\acks{
M. Haddouche and U.~\c{S}im\c{s}ekli are supported by the European Union (ERC grant DYNASTY 101039676). The French government partly funded this work under the management of Agence Nationale de la Recherche as part of the ``France 2030'' program, reference ANR-23-IACL-0008 (PR[AI]RIE-PSAI). 
B. Guedj acknowledges partial support from the French National Agency for Research, through the programme “France 2030” and PEPR IA on grant SHARP ANR-23-PEIA-0008.
We would like to thank all the reviewers for the fruitful discussion phase, which greatly enhanced this work.
}

\bibliography{main}

\begin{thebibliography}{61}
\providecommand{\natexlab}[1]{#1}
\providecommand{\url}[1]{\texttt{#1}}
\expandafter\ifx\csname urlstyle\endcsname\relax
  \providecommand{\doi}[1]{doi: #1}\else
  \providecommand{\doi}{doi: \begingroup \urlstyle{rm}\Url}\fi

\bibitem[Alquier(2024)]{alquier2021user}
Pierre Alquier.
\newblock {User-friendly Introduction to PAC-Bayes Bounds}.
\newblock \emph{Foundations and Trends\textregistered\xspace in Machine Learning}, 2024.

\bibitem[Alquier et~al.(2016)Alquier, Ridgway, and Chopin]{alquier2016properties}
Pierre Alquier, James Ridgway, and Nicolas Chopin.
\newblock {On the properties of variational approximations of Gibbs posteriors}.
\newblock \emph{Journal of Machine Learning Research}, 2016.

\bibitem[Amit et~al.(2022)Amit, Epstein, Moran, and Meir]{amit2022integral}
Ron Amit, Baruch Epstein, Shay Moran, and Ron Meir.
\newblock {Integral Probability Metrics PAC-Bayes Bounds}.
\newblock In \emph{Conference on Neural Information Processing Systems}, 2022.

\bibitem[Andriushchenko et~al.(2023)Andriushchenko, Croce, M{\"{u}}ller, Hein, and Flammarion]{andriushchenko2023modern}
Maksym Andriushchenko, Francesco Croce, Maximilian M{\"{u}}ller, Matthias Hein, and Nicolas Flammarion.
\newblock {A Modern Look at the Relationship between Sharpness and Generalization}.
\newblock In \emph{International Conference on Machine Learning}, 2023.

\bibitem[An{\'e} et~al.(2000)An{\'e}, Blach{\`e}re, Chafa{\"\i}, Foug{\`e}res, Gentil, Malrieu, Roberto, and Scheffer]{ane2000inegalites}
C{\'e}cile An{\'e}, S{\'e}bastien Blach{\`e}re, Djalil Chafa{\"\i}, Pierre Foug{\`e}res, Ivan Gentil, Florent Malrieu, Cyril Roberto, and Gr{\'e}gory Scheffer.
\newblock \emph{{Sur les in\'egalit\'es de Sobolev logarithmiques}}.
\newblock Soci\'et\'e math\'ematique de France, 2000.

\bibitem[Beckner(1989)]{beckner1989generalized}
William Beckner.
\newblock {A Generalized Poincaré Inequality for Gaussian Measures}.
\newblock \emph{Proceedings of the American Mathematical Society}, 1989.

\bibitem[Bickel and Doksum(2015)]{bickel2015mathematical}
Peter Bickel and Kjell Doksum.
\newblock \emph{{Mathematical statistics: basic ideas and selected topics}}.
\newblock CRC Press, 2015.

\bibitem[Brascamp and Lieb(1976)]{brascamp1976extensions}
Herm Brascamp and Elliott Lieb.
\newblock {On extensions of the Brunn-Minkowski and Pr\'ekopa-Leindler theorems, including inequalities for log concave functions, and with an application to the diffusion equation}.
\newblock \emph{Journal of functional analysis}, 1976.

\bibitem[Catoni(2007)]{catoni2007pac}
Olivier Catoni.
\newblock \emph{{PAC-Bayesian supervised classification: the thermodynamics of statistical learning}}.
\newblock Institute of Mathematical Statistics, 2007.

\bibitem[Chafa{\"\i}(2004)]{chafai2004entropies}
Djalil Chafa{\"\i}.
\newblock {Entropies, convexity, and functional inequalities, On $\Phi$-entropies and $\Phi$-Sobolev inequalities}.
\newblock \emph{Journal of Mathematics of Kyoto University}, 2004.

\bibitem[Chugg et~al.(2023)Chugg, Wang, and Ramdas]{chugg2023unified}
Ben Chugg, Hongjian Wang, and Aaditya Ramdas.
\newblock {A Unified Recipe for Deriving (Time-Uniform) PAC-Bayes Bounds}.
\newblock \emph{Journal of Machine Learning Research}, 2023.

\bibitem[Csisz{\'a}r(1975)]{csizar1975divergence}
I.~Csisz{\'a}r.
\newblock {$I$-Divergence Geometry of Probability Distributions and Minimization Problems}.
\newblock \emph{The Annals of Probability}, 1975.

\bibitem[Donsker and Varadhan(1976)]{donsker1976asymptotic}
M.~D. Donsker and S.~R.~S. Varadhan.
\newblock {Asymptotic evaluation of certain Markov process expectations for large time---III}.
\newblock \emph{Communications on Pure and Applied Mathematics}, 1976.

\bibitem[Dziugaite and Roy(2017)]{dziugaite2017computing}
Gintare Dziugaite and Daniel Roy.
\newblock {Computing Nonvacuous Generalization Bounds for Deep (Stochastic) Neural Networks with Many More Parameters than Training Data}.
\newblock In \emph{Conference on Uncertainty in Artificial Intelligence}, 2017.

\bibitem[Dziugaite et~al.(2020)Dziugaite, Drouin, Neal, Rajkumar, Caballero, Wang, Mitliagkas, and Roy]{dziugaite2020search}
Gintare Dziugaite, Alexandre Drouin, Brady Neal, Nitarshan Rajkumar, Ethan Caballero, Linbo Wang, Ioannis Mitliagkas, and Daniel Roy.
\newblock {In search of robust measures of generalization}.
\newblock In \emph{Conference on Neural Information Processing Systems}, 2020.

\bibitem[Foret et~al.(2021)Foret, Kleiner, Mobahi, and Neyshabur]{foret2021sharpness}
Pierre Foret, Ariel Kleiner, Hossein Mobahi, and Behnam Neyshabur.
\newblock {Sharpness-aware Minimization for Efficiently Improving Generalization}.
\newblock In \emph{International Conference on Learning Representations}, 2021.

\bibitem[Garrigos and Gower(2023)]{garrigos2023handbook}
Guillaume Garrigos and Robert~M. Gower.
\newblock {Handbook of Convergence Theorems for (Stochastic) Gradient Methods}.
\newblock \emph{arXiv}, abs/2301.11235, 2023.

\bibitem[Gastpar et~al.(2024)Gastpar, Nachum, Shafer, and Weinberger]{gastpar2023fantastic}
Michael Gastpar, Ido Nachum, Jonathan Shafer, and Thomas Weinberger.
\newblock {Fantastic Generalization Measures are Nowhere to be Found}.
\newblock In \emph{International Conference on Learning Representations}, 2024.

\bibitem[Gat et~al.(2022)Gat, Adi, Schwing, and Hazan]{gat2022importance}
Itai Gat, Yossi Adi, Alexander Schwing, and Tamir Hazan.
\newblock {On the Importance of Gradient Norm in PAC-Bayesian Bounds}.
\newblock In \emph{Conference on Neural Information Processing Systems}, 2022.

\bibitem[Ghadimi and Lan(2013)]{ghadimi2013stochastic}
Saeed Ghadimi and Guanghui Lan.
\newblock {Stochastic First- and Zeroth-order Methods for Nonconvex Stochastic Programming}.
\newblock \emph{SIAM Journal on Optimization}, 2013.

\bibitem[Glorot and Bengio(2010)]{glorot2010understanding}
Xavier Glorot and Yoshua Bengio.
\newblock {Understanding the difficulty of training deep feedforward neural networks}.
\newblock In \emph{International Conference on Artificial Intelligence and Statistics}, 2010.

\bibitem[Gross(1975)]{gross1975logarithmic}
Leonard Gross.
\newblock {Logarithmic Sobolev Inequalities}.
\newblock \emph{American Journal of Mathematics}, 1975.

\bibitem[Guedj(2019)]{guedj2019primer}
Benjamin Guedj.
\newblock {A Primer on PAC-Bayesian Learning}.
\newblock In \emph{Proceedings of the second congress of the French Mathematical Society}, 2019.

\bibitem[Guionnet and Zegarlinksi(2003)]{guionnet2003lectures}
Alice Guionnet and Bogus{\l}aw Zegarlinksi.
\newblock {Lectures on Logarithmic Sobolev Inequalities}.
\newblock \emph{S{\'e}minaire de probabilit{\'e}s XXXVI}, 2003.

\bibitem[Haddouche and Guedj(2022)]{haddouche2022online}
Maxime Haddouche and Benjamin Guedj.
\newblock {Online PAC-Bayes Learning}.
\newblock In \emph{Conference on Neural Information Processing Systems}, 2022.

\bibitem[Haddouche and Guedj(2023{\natexlab{a}})]{haddouche2023pac}
Maxime Haddouche and Benjamin Guedj.
\newblock {PAC-Bayes Generalisation Bounds for Heavy-Tailed Losses through Supermartingales}.
\newblock \emph{Transactions on Machine Learning Research}, 2023{\natexlab{a}}.

\bibitem[Haddouche and Guedj(2023{\natexlab{b}})]{haddouche2023wasserstein}
Maxime Haddouche and Benjamin Guedj.
\newblock {Wasserstein PAC-Bayes Learning: A Bridge Between Generalisation and Optimisation}.
\newblock \emph{arXiv}, abs/2304.07048, 2023{\natexlab{b}}.

\bibitem[Hellstr{\"o}m et~al.(2023)Hellstr{\"o}m, Durisi, Guedj, and Raginsky]{hellstrom2023generalization}
Fredrik Hellstr{\"o}m, Giuseppe Durisi, Benjamin Guedj, and Maxim Raginsky.
\newblock {Generalization Bounds: Perspectives from Information Theory and PAC-Bayes}.
\newblock \emph{arXiv}, abs/2309.04381, 2023.

\bibitem[Hochreiter and Schmidhuber(1997)]{hochreiter1997flat}
Sepp Hochreiter and J{\"u}rgen Schmidhuber.
\newblock {Flat minima}.
\newblock \emph{Neural computation}, 1997.

\bibitem[Jastrzebski et~al.(2017)Jastrzebski, Kenton, Arpit, Ballas, Fischer, Bengio, and Storkey]{jastrzkebski2017three}
Stanislaw Jastrzebski, Zachary Kenton, Devansh Arpit, Nicolas Ballas, Asja Fischer, Yoshua Bengio, and Amos~J. Storkey.
\newblock {Three Factors Influencing Minima in SGD}.
\newblock \emph{arXiv}, abs/1711.04623, 2017.

\bibitem[Jiang et~al.(2020)Jiang, Neyshabur, Mobahi, Krishnan, and Bengio]{jiang2020fantastic}
Yiding Jiang, Behnam Neyshabur, Hossein Mobahi, Dilip Krishnan, and Samy Bengio.
\newblock {Fantastic Generalization Measures and Where to Find Them}.
\newblock In \emph{International Conference on Learning Representations}, 2020.

\bibitem[Kuzborskij et~al.(2019)Kuzborskij, Cesa{-}Bianchi, and Szepesv{\'{a}}ri]{kuzborskij2019distribution}
Ilja Kuzborskij, Nicol{\`{o}} Cesa{-}Bianchi, and Csaba Szepesv{\'{a}}ri.
\newblock {Distribution-Dependent Analysis of Gibbs-ERM Principle}.
\newblock In \emph{Conference on Learning Theory}, 2019.

\bibitem[LeCun(1998)]{lecun1998mnist}
Yann LeCun.
\newblock {The MNIST database of handwritten digits}.
\newblock \emph{http://yann.lecun.com/exdb/mnist/}, 1998.

\bibitem[Ledoux(2006)]{ledoux2006concentration}
Michel Ledoux.
\newblock {Concentration of measure and logarithmic Sobolev inequalities}.
\newblock In \emph{S{\'e}minaire de Probabilit{\'e}s XXXIII}, 2006.

\bibitem[Li et~al.(2019)Li, Haque, Anil, Lucas, Grosse, and Jacobsen]{li2019prevent}
Qiyang Li, Saminul Haque, Cem Anil, James Lucas, Roger~B. Grosse, and J{\"{o}}rn{-}Henrik Jacobsen.
\newblock {Preventing Gradient Attenuation in Lipschitz Constrained Convolutional Networks}.
\newblock In \emph{Conference on Neural Information Processing Systems}, 2019.

\bibitem[Li and Liu(2022)]{li2022sgd}
Shaojie Li and Yong Liu.
\newblock {High Probability Guarantees for Nonconvex Stochastic Gradient Descent with Heavy Tails}.
\newblock In \emph{International Conference on Machine Learning}, 2022.

\bibitem[Lugosi and Neu(2022)]{lugosi2022generalization}
G{\'{a}}bor Lugosi and Gergely Neu.
\newblock {Generalization Bounds via Convex Analysis}.
\newblock In \emph{Conference on Learning Theory}, 2022.

\bibitem[Lugosi and Neu(2023)]{lugosi2023online}
Gábor Lugosi and Gergely Neu.
\newblock {Online-to-PAC Conversions: Generalization Bounds via Regret Analysis}.
\newblock \emph{arXiv}, abs/2305.19674, 2023.

\bibitem[Madry et~al.(2018)Madry, Makelov, Schmidt, Tsipras, and Vladu]{madry2018deep}
Aleksander Madry, Aleksandar Makelov, Ludwig Schmidt, Dimitris Tsipras, and Adrian Vladu.
\newblock {Towards Deep Learning Models Resistant to Adversarial Attacks}.
\newblock In \emph{International Conference on Learning Representations}, 2018.

\bibitem[McAllester(1999)]{mcallester1999some}
David~A. McAllester.
\newblock {Some PAC-Bayesian Theorems}.
\newblock \emph{Machine Learning}, 1999.

\bibitem[Negrea et~al.(2019)Negrea, Haghifam, Dziugaite, Khisti, and Roy]{negrea2019info}
Jeffrey Negrea, Mahdi Haghifam, Gintare~Karolina Dziugaite, Ashish Khisti, and Daniel~M. Roy.
\newblock {Information-Theoretic Generalization Bounds for SGLD via Data-Dependent Estimates}.
\newblock In \emph{Conference on Neural Information Processing Systems}, 2019.

\bibitem[Neu(2021)]{neu2021info}
Gergely Neu.
\newblock {Information-Theoretic Generalization Bounds for Stochastic Gradient Descent}.
\newblock In \emph{Conference on Learning Theory}, 2021.

\bibitem[Neyshabur et~al.(2017)Neyshabur, Bhojanapalli, McAllester, and Srebro]{neyshabur2017exploring}
Behnam Neyshabur, Srinadh Bhojanapalli, David McAllester, and Nati Srebro.
\newblock {Exploring Generalization in Deep Learning}.
\newblock In \emph{Conference on Neural Information Processing Systems}, 2017.

\bibitem[Otto and Villani(2000)]{otto2000generalization}
F.~Otto and C.~Villani.
\newblock {Generalization of an Inequality by Talagrand and Links with the Logarithmic Sobolev Inequality}.
\newblock \emph{Journal of Functional Analysis}, 2000.

\bibitem[Panageas and Piliouras(2017)]{panageas2017gradient}
Ioannis Panageas and Georgios Piliouras.
\newblock {Gradient Descent Only Converges to Minimizers: Non-Isolated Critical Points and Invariant Regions}.
\newblock In \emph{Innovations in Theoretical Computer Science Conference}, 2017.

\bibitem[P{\'{e}}rez{-}Ortiz et~al.(2021)P{\'{e}}rez{-}Ortiz, Rivasplata, Parrado{-}Hern{\'{a}}ndez, Guedj, and Shawe{-}Taylor]{perezortiz2021progress}
Mar{\'{\i}}a P{\'{e}}rez{-}Ortiz, Omar Rivasplata, Emilio Parrado{-}Hern{\'{a}}ndez, Benjamin Guedj, and John Shawe{-}Taylor.
\newblock {Progress in Self-Certified Neural Networks}.
\newblock In \emph{NeurIPS 2021 Workshop on Bayesian Deep Learning}, 2021.

\bibitem[Petzka et~al.(2021)Petzka, Kamp, Adilova, Sminchisescu, and Boley]{Petzka2021relative}
Henning Petzka, Michael Kamp, Linara Adilova, Cristian Sminchisescu, and Mario Boley.
\newblock {Relative Flatness and Generalization}.
\newblock \emph{Conference on Neural Information Processing Systems}, 2021.

\bibitem[Rivasplata et~al.(2020)Rivasplata, Kuzborskij, Szepesvari, and Shawe-Taylor]{rivasplata2020pac}
Omar Rivasplata, Ilja Kuzborskij, Csaba Szepesvari, and John Shawe-Taylor.
\newblock {PAC-Bayes Analysis Beyond the Usual Bounds}.
\newblock In \emph{Conference on Neural Information Processing Systems}, 2020.

\bibitem[Schlichting(2019)]{schlichting2019poincare}
Andr{\'{e}} Schlichting.
\newblock {Poincar\'e and Log-Sobolev Inequalities for Mixtures}.
\newblock \emph{Entropy}, 2019.

\bibitem[Shawe{-}Taylor and Williamson(1997)]{shawetaylor1997pac}
John Shawe{-}Taylor and Robert~C. Williamson.
\newblock {A PAC Analysis of a Bayesian Estimator}.
\newblock In \emph{Conference on Computational Learning Theory}, 1997.

\bibitem[Springenberg et~al.(2015)Springenberg, Dosovitskiy, Brox, and Riedmiller]{springenberg2015striving}
Jost Springenberg, Alexey Dosovitskiy, Thomas Brox, and Martin Riedmiller.
\newblock {Striving for Simplicity: The All Convolutional Net}.
\newblock In \emph{International Conference on Learning Representations -- Workshop Track}, 2015.

\bibitem[Tolstikhin and Seldin(2013)]{tolstikhin2013pac}
Ilya Tolstikhin and Yevgeny Seldin.
\newblock {PAC-Bayes-Empirical-Bernstein Inequality}.
\newblock In \emph{Conference on Neural Information Processing Systems}, 2013.

\bibitem[Viallard et~al.(2023{\natexlab{a}})Viallard, Germain, Habrard, and Morvant]{viallard2023general}
Paul Viallard, Pascal Germain, Amaury Habrard, and Emilie Morvant.
\newblock {A general framework for the practical disintegration of PAC-Bayesian bounds}.
\newblock \emph{Machine Learning}, 2023{\natexlab{a}}.

\bibitem[Viallard et~al.(2023{\natexlab{b}})Viallard, Haddouche, {\c S}im{\c s}ekli, and Guedj]{viallard2023learning}
Paul Viallard, Maxime Haddouche, Umut {\c S}im{\c s}ekli, and Benjamin Guedj.
\newblock {Learning via Wasserstein-Based High Probability Generalisation Bounds}.
\newblock In \emph{Conference on Neural Information Processing Systems}, 2023{\natexlab{b}}.

\bibitem[Viallard et~al.(2024{\natexlab{a}})Viallard, Emonet, Habrard, Morvant, and Zantedeschi]{viallard2024leveraging}
Paul Viallard, R{\'{e}}mi Emonet, Amaury Habrard, Emilie Morvant, and Valentina Zantedeschi.
\newblock {Leveraging PAC-Bayes Theory and Gibbs Distributions for Generalization Bounds with Complexity Measures}.
\newblock In \emph{International Conference on Artificial Intelligence and Statistics}, 2024{\natexlab{a}}.

\bibitem[Viallard et~al.(2024{\natexlab{b}})Viallard, Haddouche, {\c S}im{\c s}ekli, and Guedj]{viallard2024tighter}
Paul Viallard, Maxime Haddouche, Umut {\c S}im{\c s}ekli, and Benjamin Guedj.
\newblock {Tighter Generalisation Bounds via Interpolation}.
\newblock \emph{arXiv}, abs/2402.05101, 2024{\natexlab{b}}.

\bibitem[Villani(2009)]{villani2009optimal}
C{\'e}dric Villani.
\newblock \emph{{Optimal transport: old and new}}.
\newblock Springer, 2009.

\bibitem[Wen et~al.(2023)Wen, Li, and Ma]{wen2023sharpness}
Kaiyue Wen, Zhiyuan Li, and Tengyu Ma.
\newblock {Sharpness Minimization Algorithms Do Not Only Minimize Sharpness To Achieve Better Generalization}.
\newblock In \emph{Conference on Neural Information Processing Systems}, 2023.

\bibitem[Xiao et~al.(2017)Xiao, Rasul, and Vollgraf]{xiao2017fashion}
Han Xiao, Kashif Rasul, and Roland Vollgraf.
\newblock {Fashion-MNIST: a Novel Image Dataset for Benchmarking Machine Learning Algorithms}, 2017.

\bibitem[Yue et~al.(2023)Yue, Jiang, Ye, Gao, Liu, and Zhang]{yue2023sharpness}
Yun Yue, Jiadi Jiang, Zhiling Ye, Ning Gao, Yongchao Liu, and Ke~Zhang.
\newblock {Sharpness-Aware Minimization Revisited: Weighted Sharpness as a Regularization Term}.
\newblock In \emph{ACM SIGKDD Conference on Knowledge Discovery and Data Mining}, 2023.

\bibitem[Zhang(2006)]{zhang2006info}
Tong Zhang.
\newblock {Information-theoretic upper and lower bounds for statistical estimation}.
\newblock \emph{IEEE Transactions on Information Theory}, 2006.

\end{thebibliography}
\newpage
\appendix

\section{Supplementary details}
\label{sec:supp-background}
\subsection{Additional details on Poincaré and Log-Sobolev inequalities}

We first provide a proof of \Cref{prop:gibbs-logsob}.

\propgibbslogsob*

\begin{proof}
We define $\P_1$ such that $d\P_1(h)\propto \exp\left(-V(h){-}\frac{\gamma}{m}\sum_{i=1}^m \ell_1(h,\z_i)\right)dh$.
Then, by the convexity assumption over the loss $\ell_1$, we have $\Hess(V{+}\frac{\gamma}{m}\sum_{i=1}^m \ell_1(h,\z_i) \succeq \frac{1}{c_{LS}}\Id$. Then, applying \citet[Corollary 2.1]{chafai2004entropies}, we know that $\P_1$ satisfies a Poincaré inequality with constant $c_{LS}(\P)$. 
Finally, defining $\P_2$ such that $dP_2(h) \propto \exp\left(-\frac{\gamma}{m}\sum_{i=1}^m \ell_2(h,\z_i)\right)dh$, thanks to the boundedness of the loss $\ell_2$, we use \citet[Property 4.6]{guionnet2003lectures}, which ensures that $dP_2(h) = \Q^{\gamma}_{\S_m}d \P_1(h)$ satisfies a Log-Sobolev inequality with constant $c_{LS}(\P)\exp(4\|\ell_2\|_{\infty})$.
Observing that $\P_2=\Q^{\gamma}_{\S_m}$ completes the proof. 
\end{proof}

\noindent{}For the sake of completeness, we prove \Cref{prop:ls-implies-poinc}, which is Proposition 2.1 of \citet{ledoux2006concentration}, showing that a Log-Sobolev inequality implies a Poincaré inequality.

 \begin{proposition}[Proposition 2.1 of \citet{ledoux2006concentration}]
    \label{prop:ls-implies-poinc}
    If the distribution $\Q$ is $\Lsob(c_{LS})$, then it is also $\Poinc(c_\P)$, with $c_\P(\Q)= \frac{c_{LS}(\Q)}{2}$.
\end{proposition}
\begin{proof}
Let $f\in \Hrm^{1}(\Q)$ such that $\EE_{h\sim\Q}[f(h)]=0$ and $\EE_{h\sim\Q}[f^2(h)]=1$.
For any $\varepsilon>0$, we have $1+ \varepsilon f \in \Hrm^{1}(\Q)$.
We then apply the Log-Sobolev inequality on $1+\varepsilon f$ to obtain
\begin{align*}
\EE_{h\sim\Q}\left[ (1+ \varepsilon f(h))^2 \left( 2 \log(1+\varepsilon f(h)) - \log(1+\varepsilon^2)  \right)  \right] \le c_{LS}(\Q) \varepsilon^2 \EE_{h\sim\Q}\left[ \|\nabla_{h} f(h)\|^2 \right].
\end{align*}
Note that, by a Taylor expansion, we have $\log(1+ \varepsilon f(h))= \varepsilon f(h) -\frac{(\varepsilon f(h))^2}{2} + o(\varepsilon^2)$ and we have also $\log(1+ \varepsilon^2)= \varepsilon^2 + o(\varepsilon^2)$.
Then, plugging this into the previous equation gives 
\begin{align*}
      \EE_{h\sim\Q}\left[ 2\varepsilon f(h) + 3 (\varepsilon f(h))^2 - \varepsilon^2 + o(\varepsilon^2)  \right] \le c_{LS}(\Q) \varepsilon^2 \EE_{h\sim\Q}\left[ \|\nabla_{h} f(h)\|^2 \right].
\end{align*}
We use that $\EE_{h\sim\Q}[f(h)]=0$ and we then divide by $\varepsilon^2$. Taking the limit $\varepsilon \rightarrow 0$ gives: 
\begin{align*}
    \EE_{h\sim\Q}\left[ 3 f(h)^2 - 1  \right] & \le c_{LS}(\Q)\EE_{h\sim\Q}\left[ \|\nabla_{h} f(h)\|^2 \right].
\end{align*}
Using that $\EE_{h\sim\Q}[f^2(h)]=1$, we obtain
\begin{align*}
    1 \le \frac{c_{LS}(\Q)}{2}\EE_{h\sim\Q}\left[ \|\nabla_{h} f(h)\|^2 \right].
\end{align*}
Then, for any $g\in\Hrm^{1}(\Q)$ applying this proof on $f: h \mapsto \frac{g(h)- \EE_{h\sim\Q}[g(h)]}{\sqrt{\Var_{h\sim\Q}(g(h))}}$ concludes the proof.
\end{proof}

\subsection{Wasserstein distances}

We recall the definitions of the $1$-Wasserstein and the $2$-Wasserstein distances, which are valid for any predictor space $\H\subseteq \R^d$ equipped with the Euclidean distance.

\begin{definition}
\label{def:wasserstein}
The $1$-Wasserstein distance between $\Q\in\Mcal(\H)$ and $\P\in\Mcal(\H)$ is defined as
\begin{align*}
W_1(\Q,\P) = \inf_{\pi \in \Pi(\Q,\P)} \int_{\H^2} \|x-y\|d\pi(x,y).
\end{align*}
where $\Pi(\Q,\P)$ denotes the set of probability measures on $\H^2$ whose marginals are $\Q$ and $\P$.
Similarly, the $2$-Wasserstein distance between $\Q\in\Mcal(\H)$ and $\P\in\Mcal(\H)$ is defined as
\begin{align*}
W_2(\Q,\P) = \sqrt{\inf_{\pi \in \Pi(\Q,\P)} \int_{\H^2} \|x-y\|^2 d\pi(x,y)}.
\end{align*}

\end{definition}

\subsection{Fundamental difference between time-uniform PAC-Bayes bounds and classical ones}
\label{sec:fundamental-diff-time-unif}

In this work, we establish \emph{time-uniform estimation} PAC-Bayes bounds.
Specifically, we focus on the bounds such that there exist $\Ccal \subseteq \Mcal(\H)$, a threshold $\varepsilon>0$, and $\alpha>0$ such that, with probability at least $1-\delta$, for all $\Q\in \Ccal$ and $m\in\N^*$, we have
\begin{align*}
\Risk_\D(\Q) \le   \frac{\alpha}{m}\left[\KL(\Q,\P) +\log(1/\delta)\right] + \varepsilon_m.
\end{align*} 
We show that, in a favourable setting, time-uniform estimation PAC-Bayes bounds are sufficient to ensure the almost sure convergence of $(\Risk_\D(\Q_m))_{m\in\N^*}$ for a posterior sequence $(\Q_m)_{m\in\N^*}$, whereas classical PAC-Bayes bounds provide only convergence in probability.

Let us first consider a nonnegative loss $\ell$, a prior $\P$, a countable dataset $\S$, and assume there exists a sequence $(\Q_m)_{m\in\N^*}$ which is such that $\KL(\Q_m,\P)\le D$ for all $m\in\N^*$ and both $\hat{R}_{\S_m}(\Q_m)$ and $\EE_{\z\sim\D,h\sim \Q_m}\left[\|\nabla_h \ell(h,\z)\|^2\right]$ go to zero as $m$ goes to infinity almost surely. 
Then, the classical McAllester's bound states that for any $m\in\N^*$, and $\delta_m\in(0,1]$, with probability at least $1-\delta_m$ over $\S_m$, we have 
\begin{align*}
{\textstyle\PP_{\S_m}}\left( \Risk_{\D}(\Q_m)   \le \hat{\Risk}_{\S_m}(\Q_m) + \sqrt{\frac{\KL(\Q_m,\P) +\log(2\sqrt{m}/\delta_m)}{2m}}\right) \geq 1- \delta_m.
\end{align*}
Let $\alpha>0$ and take for all $m\in\N^*$, the confidence parameter $\delta_m= \delta/m$. Since we assume that $\hat{\Risk}_{\S_m}(\Q_m) \rightarrow 0$, the square root in McAllester's bound goes to zero. Thus, there exists $m_0\in\N^*$ such that for all $m\geq m_0$, we have  
\begin{align*}
\Risk_{\D}(\Q_m)   \le \hat{\Risk}_{\S_m}(\Q_m) + \sqrt{\frac{\KL(\Q_m,\P) +\log(2m\sqrt{m}/\delta)}{2m}}  \le \alpha.
\end{align*}
Thus, for all $m \geq m_0$, we know that $\PP_{\S_m}\left(\Risk_{\D}(\Q_m) < \alpha  \right) \geq 1-\delta/m.$
Taking the limit as $m$ goes to infinity ensures that for any $\alpha>0$, we have 
\begin{align*}
\lim_{m\rightarrow \infty} {\textstyle\PP_{\S_m}}\left(\Risk_{\D}(\Q_m) < \alpha  \right) = 1.
\end{align*}
Thus, the classical McAllester's bound allows the sequence $\Risk_{\D}(\Q_m)$ to converge in probability to zero. 
In contrast, \Cref{th:poincare-gauss} with $\lambda = 1/C$ is a time-uniform estimation bound with $\alpha= 2C$ and $\varepsilon_m = 2\left(\hat{R}_{\S_m}(\Q) + \frac{1}{2C} \EE_{\z\sim\D,h\sim\Q}\left[\|\nabla_h \ell(h,\z)\|^2\right]\right)$.
Under our assumptions, we know that both $\frac{\alpha}{m}\left[\KL(\Q,\P) +\log(1/\delta)\right]$ and $\varepsilon_m$ go to zero as $m$ goes to infinity.
Then, taking the limit, we have for all $\delta>0$
\begin{align*}
{\textstyle\PP_{\S}}\left(\lim_{m\rightarrow\infty}\Risk_{\D}(\Q_m) =0  \right) \geq 1-\delta.
\end{align*}
As $\lim_{m\rightarrow\infty}\Risk_{\D}(\Q_m)$ does not depend on $\delta$, we can then make $\delta$ go to zero to obtain 
\begin{align*}
{\textstyle\PP_{\S}}\left(\lim_{m\rightarrow\infty}\Risk_{\D}(\Q_m) =0  \right)=1.
\end{align*}
Thus, $\Risk_{\D}(\Q_m)$ converges almost surely to zero. 
Such a conclusion cannot be achieved with classical PAC-Bayes bounds; our bound is stronger as it allows an almost sure convergence, which cannot be obtained by simple manipulations of classical bounds.

\subsection{A deeper look at the nature of our gradient terms and their originality in PAC-Bayes}
\label{sec:comparison-tolstikhin}

Here, we highlight the novelty of incorporating $\nabla_h \ell(h,\z)$ in PAC-Bayes bounds and argue that this is not a trivial extension of the PAC-Bayes Bernstein bound.
Given a bounded loss $\ell:\H \times \Zcal \to [0,1]$, and starting from the Bernstein bound of \citet{tolstikhin2013pac}, it is possible to derive a bound involving gradient terms with an additional cost when the posterior satisfies a Poincaré inequality. We first restate the PAC-Bayes Bernstein bound of \citet{tolstikhin2013pac}.

\begin{theorem}\label{th:tolstikhin}
For any $c_1>1$, for any data-free prior $\P\in\Mcal(\H)$, for any loss function $\ell: \H\times\Zcal\to[0,1]$, and for any $\delta\in (0,1]$, with probability at least $1-\delta$ over the sample $\S$, we have for all distributions $\Q\in\Mcal(\H)$ that satisfy
\begin{align}\label{eq:tolstikhin}
\sqrt{\frac{\KL(\Q, \P)+\ln \frac{\nu_1}{\delta_1}}{(e-2) \EE_{h\sim\Q}[\Var_{\z\sim\D}(\ell(h,\z))]}} \le \sqrt{m}, 
\end{align}
we have
\begin{align*}
\Risk_\D(\Q) \le \hat{\Risk}_{\S_m}(\Q)+\left(1+c_1\right) \sqrt{\frac{(e-2) \EE_{h\sim\Q}[\Var_{\z\sim\D}(\ell(h,\z))]\left(\KL(\Q, \P)+\ln \frac{\nu_1}{\delta}\right)}{m}},
\end{align*}
where
\begin{align*}
\nu_1=\left[\frac{1}{\ln c_1} \ln \left(\sqrt{\frac{(e-2) m}{4 \ln \left(1 / \delta\right)}}\right)\right]+1.
\end{align*}
\end{theorem}
Assuming the technical conditions of \Cref{th:tolstikhin}, that the distribution $\D$ satisfies a Poincaré inequality, and that the loss $\ell(h,\cdot)\in \Hrm^1(\D)$ for all $h\in\H$, we obtain the following corollary.

\begin{corollary} Under the same conditions of \Cref{th:tolstikhin}, for any distribution $\D$  being $\Poinc(c_\D)$, and for any loss $\ell(h,\cdot)\in \Hrm^1(\D)$, with probability at least $1-\delta$ over the sample $\S$, we have for all distributions $\Q\in\Mcal(\H)$ satisfying \Cref{eq:tolstikhin}
\begin{align*}
\Risk_\D(\Q) \le \hat{\Risk}_{\S_m}(\Q)+\left(1+c_1\right) \sqrt{\frac{(e-2) \EE_{h\sim\Q}\EE_{\z\sim \D}\left[ \|\nabla_{\z} \ell(h,\z)\|^2\right]\left(\KL(\Q, \P)+\ln \frac{\nu_1}{\delta}\right)}{m}}.
\end{align*}
\end{corollary}
\begin{proof}
From Poincaré inequality (\Cref{def:poincare}), we have
\begin{align*}
\EE_{h\sim\Q}\left[\Var_{\z\sim\D}(\ell(h,\z))\right] \le \EE_{h\sim\Q}\EE_{\z\sim \D}\left[ \|\nabla_{\z}\ell(h,\z)\|^2\right],
\end{align*}
which is then substituted into the bound of \Cref{th:tolstikhin}. 
\end{proof}
By using the additional Poincaré assumption on the data distribution $\D$, we can obtain a bound with gradient term $\EE_{h\sim\Q}\EE_{\z\sim \D}[ \|\nabla_{\z} \ell(h,\z)\|^2]$.
However, it remains unclear how $\nabla_\z\ell(h,\z)$ behaves, as we do not optimise with respect to $\z$.
As a result, these gradients may remain large even after a successful learning phase.
In contrast, \Cref{th:poincare-gauss} involves the term $\EE_{\z\sim\D}\EE_{h\sim \Q}[ \|\nabla_h \ell(h,\z)\|^2]$, which is minimised during a successful optimisation process.
Moreover, the Poincaré assumption on the data distribution $\D$ is difficult to verify, as we do not know $\D$. Our Poincaré assumption on the posterior $\Q$ is much easier to verify, as we can choose $\Q$ in practice.
A similar discussion then applies for the empirical Bernstein bound \citep[Theorem 4]{tolstikhin2013pac} as the empirical variance is with respect to the dataset $\S$.

\subsection{How to make the Wasserstein term go to zero in \Cref{th:wpb-grad}?}
\label{sec:wass-to-zero}

In this section, we prove the following lemma.

\begin{lemma}
Assume that the hypothesis set $\H$ is finite and that the loss $\ell: \H\times \Zcal \to \Rbb$ satisfies $(\star)$ with constant $G$.
Then, with probability at least $1-\delta$, the generalisation gap $\Risk_\D(h)-\hat{\Risk}_{\S_m}(h)$ satisfies $(\star)$ with the constant $G'=\Ocal\left( G\sqrt{\nicefrac{\log(|\H|/\delta)}{m}}\right)$.
\end{lemma}

\begin{proof}
For the sake of simplicity, let $f :h\mapsto\Risk_\D(h)-\hat{\Risk}_{\S_m}(h)$ be the generalisation gap.
Then, fix $(h_1,h_2,a,a')\in\H^4$ and our goal is to prove that $f$ satisfies $(\star)$ with another constant $G'$.
First of all, notice that
\begin{align*}
    &\left\langle\nabla_{h} f(h_1)-\nabla_{h} f(h_2), a-a' \right\rangle \\
    &= \frac{1}{m} \sum_{i=1}^m  \left\langle\nabla_{h}\ell(h_2,\z_i) - \nabla_{h}\ell(h_1,\z_i) -(\nabla_{h} \Risk_{\D}(h) -\nabla_{h} \Risk_{\D}(h)),a -a'\right\rangle.
\end{align*}
Moreover, by the condition $(\star)$ for all $\z\in\Zcal$, we know that 
\begin{align*}
|\langle \nabla_{h}\ell(h_2,\z) - \nabla_{h}\ell(h_1,\z),a - a' \rangle| \le G\cdot\|a-a'\|\cdot\|h_1-h_2\|.
\end{align*}
\looseness=-1
Then, by Hoeffding's inequality, applied on the centered random variable $\nabla_{h}\ell(h_2,\z_i) - \nabla_{h}\ell(h_1,\z_i) -(\nabla_{h} \Risk_{\D}(h_2) -\nabla_{h} \Risk_{\D}(h_1))$ bounded by $ G\|a-a'\|\|h_1-h_2\|$, with probability at least $1-\delta$, we have
\begin{align*}
\left|\left\langle\nabla_{h} f(h_1)-\nabla_{h} f(h_2), a - a' \right\rangle \right| \le G\cdot\|a-a'\|\cdot\|h_1-h_2\|\cdot\sqrt{\frac{2\log(2/\delta)}{m}}.
\end{align*}
Taking a union bound on all possible values of $(h_1,h_2,a,a')\in\H^4$ with $\delta'= \delta/|\H|^4$ and a union bound on all tuples yields that, with probability at least $1-\delta$, for all $(a,a')\in\H^2$, 
\begin{align*}
\text{the function } h\mapsto \langle \nabla_h f(h),a-a'\rangle \text{ is } G\sqrt{\frac{2\log\left(\frac{2|\H|^4}{\delta}\right)}{m}}\|a-a'\|\text{-Lipschitz,}
\end{align*}
meaning the condition $(\star)$ is verified with constant $G'= G\sqrt{\frac{2\log\left(\frac{2|\H|^4}{\delta}\right)}{m}}$ for the gap.
\end{proof}

\section{PAC-Bayes bounds for Lipschitz losses through log-Sobolev inequalities}
\label{sec:lpz-results}

\paragraph{Extending Catoni's bound to Lipschitz losses.}
A well-known relaxation of \citet[Theorem 1.2.6]{catoni2007pac} \citep[see also][Theorem 4.1]{alquier2016properties} holding for subgaussian losses has been widely used in practice as a tractable PAC-Bayesian algorithm exhibiting a linear dependency on the KL divergence.
Below, we exploit a consequence of the Herbst argument, as stated, for example, in \citet[Section 2.3]{ledoux2006concentration}, which asserts that an $L$-Lipschitz function of a random variable following a distribution $\D$ being $\Lsob(c_{LS})$ is $L\sqrt{c_{LS}(\D)}$-subgaussian. This leads to the following corollary. 

\begin{corollary}\label{cor:catoni-lpz}
For any $\lambda >0$, for any data-free prior $\P\in\Mcal(\H)$, for any $L$-Lipschitz loss $\ell : \H\times \Zcal\to \R$ for any $h\in\H$, for any data distribution $\D$ being $\Lsob(c_{LS})$, with probability at least $1-\delta$ over $\S$, for any $\Q\in \Mcal(\H)$, we have
\begin{align*}
\Risk_\D(\Q) \le \hat{\Risk}_{\S_m}(\Q) + \frac{\KL(\Q,\P) + \log(1/\delta)}{\lambda} + \frac{\lambda^2L^2c_{LS}(\D)}{2m}.
\end{align*}
\end{corollary}

\begin{proof}
First, we take $f(h,\S_m)\defeq\lambda (\Risk_\D(h) - \hat{\Risk}_{\S_m}(h))$ and we use the change of measure inequality \citep{csizar1975divergence,donsker1976asymptotic} to state that, for all $\Q\in\Mcal(\H)$, we have 
\begin{align*}
\EE_{h\sim\Q}[f(h, \S_m)] \le \KL(\Q,\P) + \log \left(\EE_{h\sim\P} \left[ \exp \left( f(h, \S_m) \right) \right] \right).
\end{align*}
Moreover, Markov's inequality alongside Fubini's theorem gives, with probability at least $1-\delta$ over the sample $\S$,
\begin{align*}
\EE_{h\sim\Q}[f(h, \S_m)] \le \KL(\Q,\P) + \log(1/\delta) + \log \left(\EE_{h\sim\P} \EE_{\S_m} \left[ \exp \left( f(h, \S_m) \right) \right] \right).
\end{align*}
Now, since the loss $\ell$ is $L$-Lipschitz on $\z\in\Zcal$ for all $h\in\H$, we show below that the function $f$ is $\frac{\lambda L}{\sqrt{m}}$-Lipschitz on the variable $\S_m$ for each $h\in\H$.
Indeed, as the loss is $L$-Lipschitz w.r.t. $\z\in\Zcal$, for any dataset $\S_m = (\z_1,\dots,\z_m)$, any $\S'_m = (\z'_1,\cdots,\z'_m)$, and any $h\in \H$, we have
\begin{align*}
\|f(h, \S_m) - f(h, \S_m')\| \le \frac{\lambda L}{m} \sum_{i=1}^m \|\z_i-\z'_i\|= \frac{\lambda L}{m} \sum_{i=1}^m \sqrt{\|\z_i-\z'_i\|^2}.
\end{align*}
Then, by the concavity of the square root, we have 
\begin{align*}
\|f(h, \S_m) - f(h, \S_m')\| \le \lambda L\sqrt{\frac{1}{m}\sum_{i=1}^m \|\z_i-\z'_i\|^2} = \lambda L\frac{\|\S_m - \S'_m\|}{\sqrt{m}}.
\end{align*}
The underlying norm $\|\S'_m\|$ is the one derived from the scalar product $\langle \S_m, \S'_m\rangle= \sum_{i=1}^m \langle \z_i,\z'_i\rangle$.
As the distribution $\D$ is $\Lsob(c_{LS})$, we can deduce that $\D^{\otimes m}$ is also $\Lsob(c_{LS})$ with the same constant \citep[Corollary 3.2.3]{ane2000inegalites}.
Then, using the Herbst argument similarly to \citet[Section 2.3]{ledoux2006concentration}, we conclude that for all $h\in\H$, the function $f(h,\cdot)$ is $L\lambda\sqrt{\frac{c_{LS}(\D)}{m}}$-subgaussian. 
Thus, we have
\begin{align*}
 \log \left(\EE_{h\sim\P} \EE_{\S_m} \left[ \exp \left( f(h, \S_m) \right) \right] \right) \le \frac{\lambda^2L^2c_{LS}(\D)}{2m},
\end{align*}
which concludes the proof. 
\end{proof}

\noindent{}\textbf{Disintegrated PAC-Bayes bounds.} 
Numerical estimation of PAC-Bayes bounds is usually challenging as it often involves Monte-Carlo approximations of the expectation over the posterior $\Q$.
A recent line of work \citep{rivasplata2020pac,haddouche2022online,viallard2023general,viallard2024leveraging} studies \emph{disintegrated PAC-Bayes bounds}, \ie bounds that hold with high probability on both the dataset $\S$ and a single predictor $h$ drawn from the posterior $\Q$. These bounds may be relevant for practitioners when sampling is easy, as in the case of Gaussian distributions, since they require little computational time. However, a drawback of these bounds is that they do not allow the KL divergence to be used as a complexity measure.
Instead, either the disintegrated KL \citep{rivasplata2020pac} or the Rényi divergence \citep{viallard2023general} is considered, which can be seen as a relaxation of the KL divergence one.
By leveraging the subgaussianity behaviour of Lipschitz losses, it is possible to derive PAC-Bayesian disintegrated bounds, as long as the posterior distribution satisfies a log-Sobolev inequality with a sharp constant (which can be achieved, for instance, for Gaussian distributions with a small operator norm). 
The new disintegrated bound is introduced in the following lemma. 

\begin{lemma}\label{lemma:disintegrated}
For any $L$-Lipschitz loss $\ell : \H\times \Zcal\to \R$, for any distribution $\Q$ being $\Lsob(c_{LS})$ with $c_{LS}(\Q)\le 1/m$ (and that can depend on the dataset $\S_m$), with probability $1-\delta$ over the draw of $h\sim\Q$ and $\S_m$, we have 
\begin{align*}
\Risk_\D(h){-}\hat{\Risk}_{\S_m}(h) \le \Risk_\D(\Q){-}\hat{\Risk}_{\S_m}(\Q) + \sqrt{\frac{L^2\log(1/\delta)}{2m}}.
\end{align*}
\end{lemma}
\begin{proof}
We simply remark, by the same argument as in the proof of \Cref{cor:catoni-lpz} that the gap $h \mapsto \Risk_\D(h)-\Risk_{\S_m}(h)$ is $L$-Lipschitz for any $\S_m$ thus the gap is $L\sqrt{c_{LS}(\Q)}$-subgaussian. Then we use 
\begin{align*}
\Risk_\D(h){-}\hat{\Risk}_{\S_m}(h) = \log\left(\exp\left( \Risk_\D(h){-}\hat{\Risk}_{\S_m}(h) \right) \right).
\end{align*}
We then apply Markov inequality and exploit the subgaussiannity of the gap alongside $c_{LS}(\Q)\le 1/m$ to conclude the proof.

\end{proof}
This lemma states that, as long as we assume our loss to be Lipschitz with respect to $h$, it is possible to easily derive disintegrated PAC-Bayesian bounds.
Additionally, \Cref{lemma:disintegrated} can be easily completed by \Cref{cor:catoni-lpz}, which introduces a KL divergence as a complexity term. Note also that, since the loss is Lipschitz, it is also possible to incorporate the $1$-Wasserstein distance through the bounds of \citet{haddouche2023wasserstein,viallard2023learning,viallard2024tighter}. Therefore, having a Log-Sobolev assumption with a sharp constant on the posterior distribution is enough to provide disintegrated PAC-Bayesian bounds involving the KL divergence or the Wasserstein distance, rather than the Rényi divergence or the disintegrated KL divergence.

\section{Proofs}
\label{sec:proofs}
\subsection{Proof of \Cref{cor:poincare-pacb}}\label{sec:proof-poincare-pacb}

The goal of this section is to prove \Cref{cor:poincare-pacb}, which is restated for ease of readability.

\corpoincarepacb*

\noindent{}To begin this proof, we first state an important intermediate theorem, which holds without any assumptions on the data distribution.

\begin{theorem}\label{th:poincare-pacb}
For any $C>0$, for any $\lambda$ such that $\frac{2}{C}{>}\lambda{>}0$, for any data-free prior $\P\in\Mcal(\H)$, for any loss function $\ell: \H\times\Zcal\to \R_{+}$, and for any $\delta\in (0,1]$, with probability at least $1-\delta$ over the sample $\S$, for all $m\in\N^{*}$, for all posterior $\Q$ being $\Poinc(c_\P)$ with $\Risk_\D(\Q)\le C$ and such that for any $\z\in\Zcal$, $\ell(\cdot,\z)\in \Hrm^{1}(\Q)$:
    \begin{multline*}
      \Risk_{\D}(\Q) \le \frac{1}{1-\frac{\lambda C}{2}} \left( \hat{\Risk}_{\S_m}(\Q) + \frac{\KL(\Q,\P) +\log(1/\delta)}{\lambda m} \right) \\
       + \frac{\lambda}{2-\lambda C}\left( c_{P}(\Q)\EE_{\z\sim\D} \left[ \EE_{h\sim \Q}\left( \|\nabla_h \ell(h,\z)\|^2 \right) \right]  + \Var_{\z\sim\D} \left( \EE_{h\sim\Q}[\ell(h,\z)] \right)  \right). 
    \end{multline*} 
  \end{theorem}
\Cref{th:poincare-pacb} highlights the influence of the gradient norm of $\nabla_h\ell(h,\z)$ on the generalisation ability: small gradients make the bound vanish.
The remaining variance term is not addressed at this stage and can be assumed to be bounded, but we cannot then recover a fast rate.\\ 
 
\begin{proof}
We start from \citet[Corollary 17]{chugg2023unified}, for any $\lambda >0$, with probability at least $1-\delta$, for all $m\in\N^{*}$, for all posteriors $\Q\in\H$, we have
    \begin{align*}
      \Risk_{\D}(\Q) \le  \hat{\Risk}_{\S_m}(\Q) + \frac{\KL(\Q,\P) +\log(1/\delta)}{\lambda m} 
      + \frac{\lambda }{2}\left(   \EE_{h\sim \Q}\left[\EE_{\z\sim \D}[\ell (h,\z)^2]  \right]  \right).
    \end{align*}
    The last term is then controlled as follows:
    \begin{align*}
      \EE_{h\sim \Q}\left[\EE_{\z\sim \D}[\ell (h,\z)^2] \right] &  \le  \EE_{\z\sim\D} \left[ c_{P}(\Q)\EE_{h\sim \Q}\left( \|\nabla_h \ell(h,\z)\|^2 \right) + \left( \EE_{h\sim\Q}[\ell(h,\z)] \right)^2 \right] .
      \intertext{We then introduce a supplementary variance term:}
      & =  \EE_{\z\sim\D} \left[ c_{P}(\Q)\EE_{h\sim \Q}\left( \|\nabla_h \ell(h,\z)\|^2 \right) \right]  + \Var_{\z\sim\D} \left( \EE_{h\sim\Q}[\ell(h,\z)] \right) \\
      & \hspace{4mm}+ \left( \EE_{\z\sim\D}\EE_{h\sim\Q}[\ell(h,\z)] \right)^2.
    \end{align*}
Note that by Fubini's theorem, the last term on the right-hand side is exactly $ \Risk_\D(\Q)^2$. 
Then, using the fact that the averaged true risk is less than $C$, and reorganising the terms in \citet[Corollary 17]{chugg2023unified}, we obtain, for $\lambda \in \left( 0, \frac{2}{C}\right)$:
    \begin{multline*}
      \Risk_{\D}(\Q) \le \frac{1}{1-\frac{\lambda C}{2}} \hat{\Risk}_{\S_m}(\Q) + \frac{\KL(\Q,\P) +\log(1/\delta)}{\lambda \left( 1-\frac{\lambda C}{2}\right) m} \\
      + \frac{\lambda}{2-\lambda C}\left( c_{P}(\Q)\EE_{\z\sim\D} \left[ \EE_{h\sim \Q}\left( \|\nabla_h \ell(h,\z)\|^2 \right) \right]  + \Var_{\z\sim\D} \left( \EE_{h\sim\Q}[\ell(h,\z)] \right)  \right).
    \end{multline*}
  \end{proof}

\noindent{}Now that \Cref{th:poincare-pacb} is proven, we only need to apply the Poincaré assumption on the data distribution to the variance term to derive \Cref{cor:poincare-pacb}.\\

\subsection{Proof of \Cref{th:poincare-grad-lpz}}
\label{sec:proof-poincare-grad}

\thpoincaregradlpz*

\begin{proof}
  We start again from \Cref{th:poincare-gauss}, with $\lambda= 1/C_1$, to obtain, with probability at least $1-\delta/2$:
  \begin{multline}
    \label{eq:fast-rate-grad-lpz-1}
    \Risk_{\D}(\Q) \le 2 \left( \hat{\Risk}_{\S_m}(\Q) + 2C_1\frac{\KL(\Q,\P) +\log(2/\delta)}{ m} \right) + \frac{c_\P(\Q)}{C_1}\EE_{\z\sim\D} \left[ \EE_{h\sim \Q}\left( \|\nabla_h \ell(h,\z)\|^2 \right) \right] .
  \end{multline}
  We now observe that $g: h,\z \mapsto \|\nabla_h \ell(h,\z)\|^2$ is nonnegative.
  Given our assumptions, we apply the proof technique of \Cref{th:poincare-gauss} on $g$, \ie we start again from Corollary 17 of \citep{chugg2023unified}, apply Poincaré inequality on $\Q$ and use the $\texttt{QSB}$ assumption on $g$. We then have, for any $\lambda >0$, with probability at least $1-\delta/2$, for all $\Q$ being $\Poinc(c_\P)$, $\texttt{QSB}\left( g,C_2\right)$ and $g(\cdot,\z)\in \Hrm^{1}(\Q)$ for all $\z\in\Zcal$ :
   \begin{multline}
    \label{eq:fast-rate-grad-lpz-2}
    \EE_{\z\sim\D} \left[ \EE_{h\sim \Q}\left( \|\nabla_h \ell(h,\z)\|^2 \right) \right] \le \EE_{h\sim \Q}\left[ \frac{1}{m}\sum_{i=1}^m \|\nabla_h\ell(h,\z_i)\|^2 \right]  + \frac{\KL(\Q,\P) +\log(2/\delta)}{\lambda m} \\
    + \frac{\lambda c_\P(\Q)}{2}\EE_{\z\sim\D} \left[ \EE_{h\sim \Q}\left( \|\nabla_h g(h,\z)\|^2 \right) \right] + \frac{\lambda C_2}{2}  \EE_{\z\sim\D} \left[ \EE_{h\sim \Q}\left( \|\nabla_h \ell(h,\z)\|^2 \right) \right].
  \end{multline}
  Notice that by definition of $g(\cdot,\z): \R^d \rightarrow \R$, we have $\nabla_h g(h,\z) = 2\Hess_h(\ell)(h,\z)\nabla_h \ell (h,\z)$, where $\Hess_h(\ell)$ denotes the Hessian of $\ell$. Thus, using the fact that $\ell(\cdot, \z)$ is $G$ gradient-Lipschitz for any $\z\in\Zcal$, we get, for any $(h,\z)$, that $\|\nabla_h g(h,\z)\| \le 2G \|\nabla_h \ell(h,\z)\|$. Substituting this in \Cref{eq:fast-rate-grad-lpz-2} gives: 
  \begin{multline}
    \label{eq:fast-rate-grad-lpz-3}
    \EE_{\z\sim\D} \left[ \EE_{h\sim \Q}\left( \|\nabla_h \ell(h,\z)\|^2 \right) \right] \le \EE_{h\sim \Q}\left[ \frac{1}{m}\sum_{i=1}^m \|\nabla_h\ell(h,\z_i)\|^2 \right]  + \frac{\KL(\Q,\P) +\log(2/\delta)}{\lambda m} \\
    + \frac{\lambda}{2}\left( 4c_\P(\Q)G^2 + C_2 \right)  \EE_{\z\sim\D} \left[ \EE_{h\sim \Q}\left( \|\nabla_h \ell(h,\z)\|^2 \right) \right].
  \end{multline}
  Using that $c_\P(\Q)=c$, taking $\lambda = \frac{1}{4cG^2 + C_2}$ and reorganising the terms in \Cref{eq:fast-rate-grad-lpz-3} gives:
  \begin{multline}
    \label{eq:fast-rate-grad-lpz-4}
    \EE_{\z\sim\D} \left[ \EE_{h\sim \Q}\left( \|\nabla_h \ell(h,\z)\|^2 \right) \right] \le 2\EE_{h\sim \Q}\left[ \frac{1}{m}\sum_{i=1}^m \|\nabla_h\ell(h,\z_i)\|^2 \right]  \\+ 2(4cG^2 + C_2)\frac{\KL(\Q,\P) +\log(2/\delta)}{m}. 
  \end{multline}
  Finally, taking a union bound and plugging \Cref{eq:fast-rate-grad-lpz-4} in \Cref{eq:fast-rate-grad-lpz-1} concludes the proof.
\end{proof}

\subsection{Proof of \Cref{lemma:kl-bound}}\label{sec:proof-kl-bound}

\lemmaklbound*
    \begin{proof}
  For conciseness, we rename $\Q\defeq\Q^{\gamma}_{\S_m}$. We first note that we have
  \begin{align*}
    \KL\left( \Q^{\gamma}_{\S_m},\P\right)  & = \EE_{h\sim \Q} \left[\log\left(\frac{d\Q}{d\P}(h)\right) \right]  \\
    & = \Ent_{\P} \left(\frac{d\Q}{d\P} \right) = \Ent_P[g^2],
  \end{align*}
where $g =\sqrt{\frac{d\Q}{d\P}}$ and $\frac{d\Q}{d\P}$ is the Radon-Nikodym derivative of $\Q$ with respect to $\P$.
Recall that $\frac{d\Q}{d\P}(h) = \frac{1}{Z}\exp(-\gamma \hat{\Risk}_{\S_m}(h))$, where $Z= \EE_{h\sim \P}[ \exp( -\gamma \hat{\Risk}_{\S_m}(h)) ]$.
Then, the function $g: h \mapsto \frac{1}{\sqrt{Z}}\exp(-\frac{\gamma}{2} \hat{\Risk}_{\S_m}(h))$ belongs to $\Hrm^{1}(\P)$ as long as $\ell \in \Hrm^{1}(\P)$. Indeed, since $\exp$ is infinitely smooth, $g\in \Drm_1(\R^d)$.
Also, as the loss is nonnegative, we have $g \le \frac{1}{\sqrt{Z}}$, so $g\in \Lrm^2(\P)$.
Finally, we have $\nabla_{h} g(h) = -\frac{\gamma }{2}g(h)\nabla_h \hat{\Risk}_{\S_m}(h) $. Since $g(h)\le \frac{1}{\sqrt{K}}$, we only need to bound $ \|\nabla_{h} \hat{\Risk}_{\S_m}(h)\|^2$ to ensure that $g\in \Hrm^{1}(\P)$: 
  \begin{align*}
    \|\nabla_{h} \hat{\Risk}_{\S_m}(h)\|^2 & =\frac{1}{m^2}\sum_{1\le i,j\le m}\left\langle \nabla_{h}\ell(h,\z_i), \nabla_h \ell(h,\z_j) \right\rangle \\
    & \le \frac{1}{2m^2}\sum_{1\le i,j\le m} \|\nabla_{h}\ell(h,\z_i)\|^2 + \|\nabla_{h}\ell(h,\z_j)\|^2.
  \end{align*}
  Since we assumed $\|\nabla_{h}\ell(h,\z)\|^2\in \Lrm^2(\P)$ for all $\z\in\Zcal$, we conclude that $g\in \Hrm^{1}(\P)$.
  We then can apply the log-Sobolev inequality to conclude that
  \begin{align*}
    \KL\left( \Q^{\gamma}_{\S_m},\P\right)  & \le c_{LS}(\P) \EE_{h\sim\P}[\|\nabla_{h} g(h)\|^2] \\
    & = \frac{\gamma^2 c_{LS}(\P)}{4}\EE_{h\sim\P} \left[ \|\nabla_h \hat{\Risk}_{\S_m}(h) \|^2 g^2(h) \right] \\
    &= \frac{\gamma^2 c_{LS}(\P)}{4}\EE_{h\sim\P} \left[ \|\nabla_h \hat{\Risk}_{\S_m}(h) \|^2 \frac{d\Q}{d\P}(h) \right]\\
    & =\frac{\gamma^2 c_{LS}(\P)}{4} \EE_{h\sim \Q^{\gamma}_{\S_m}}\left[ \|\nabla_h \hat{\Risk}_{\S_m}(h) \|^2 \right].
  \end{align*}
\end{proof}

\subsection{Proof of \Cref{th:gibbs-pacb}}\label{sec:proof-gibbs-pacb}
  
\thgibbspacb*  
  \begin{proof}
  We start again from \citet[Corollary 17]{chugg2023unified}, instantiated with a single $\lambda$. Then with probability at least $1-\delta$, for all posteriors $\Q$ and for all $m\in\N^{*}$, we have
  \begin{align*}
    \Risk_{\D}(\Q) \le  \hat{\Risk}_{\S_m}(\Q) + \frac{\KL(\Q,\P) +\log(1/\delta)}{\lambda m} 
    + \frac{\lambda }{2}\left(   \EE_{h\sim \Q}\left[\EE_{\z\sim \D}[\ell (h,\z)^2]  \right]  \right).
  \end{align*} 
For the first inequality, we simply take $\lambda =1$, use the fact that $\ell(h,\z)^2 \le \ell(h,\z)$, and reorganise the terms. Finally, we upper-bound the KL term using \Cref{lemma:kl-bound}.
  For the second inequality, we apply \Cref{prop:gibbs-logsob} to use the fact that $\Q^{\gamma}_{\S_m}$ is $\Lsob(c_{LS})$, and \Cref{prop:ls-implies-poinc}, which ensures that $\Q^{\gamma}_{\S_m}$ is $\Poinc(c_\P)$, with constant equal to $c_{LS}\left( \Q^{\gamma}_{\S_m}\right)/2$.
  We then follow a proof technique similar to \Cref{th:poincare-gauss}. We have : 
  \begin{align*}
    \EE_{h\sim \Q^{\gamma}_{\S_m}}\left[\EE_{\z\sim \D}[\ell (h,\z)^2] \right] &  = \EE_{\z\sim\D} \left[ \Var_{h\sim \Q^{\gamma}_{\S_m}}\left( \ell(h,\z) \right) + \left( \EE_{h\sim\Q^{\gamma}_{\S_m}}[\ell(h,\z)] \right)^2 \right].
\end{align*}
Applying Poincaré inequality then gives:
\begin{align*}
    \le  \EE_{\z\sim\D} \left[ \frac{c_{LS}(\P)}{2}e^{4\|\ell_2\|_{\infty}}\EE_{h\sim \Q^{\gamma}_{\S_m}}\left( \|\nabla_h \ell(h,\z)\|^2 \right) + \left( \EE_{h\sim\Q^{\gamma}_{\S_m}}[\ell(h,\z)] \right)^2 \right].
  \end{align*}
  Finally, using the fact that $\Q^{\gamma}_{\S_m}$ is $\texttt{QSB}(\ell,C)$ allow us to reorganise the terms as in \Cref{th:poincare-gauss}. Combining this with \Cref{lemma:kl-bound} to bound the KL divergence concludes the proof.
\end{proof}

\subsection{Proof of \Cref{th:2-wass}}
\label{sec:proof-2-wass}

\thtwowass*
\begin{proof}
  We first assume that $G=1$ in the $(\star)$ assumption. We start from the Kantorovich duality formula \citep[Theorem 5.10]{villani2009optimal}, instantiated with the cost function $c(x,y)= \|x-y\|^2$. For any $\Q,\P$, since $W_2$ is a distance, we have:
 \begin{align}
    \label{eq:kantorovich}
    W^2(\Q,\P) = W^2(\P,\Q) = \sup_{\phi, \psi}\;\EE_{h\sim\Q}[\phi(h)] -\EE_{h\sim\P}[\psi(h)], 
 \end{align}
  where the supremum is taken over the functions $\phi,\psi\in L^1(\Q)\times L^1(\P)$ such that for all $h,h' \in\H^2$, we have $\phi(h)-\psi(h')\le \|h-h'\|^2$. 
  We claim that if $\phi(h)= f(h) - D\|\nabla f(h)\|$ and $\psi(h') = f(h')$, then the pair $(\Phi,\Psi)$ satisfies $\phi(h)-\psi(h')\le \frac{\|h-h'\|^2}{2}$. 
  Indeed, we have
  \begin{align*}
    \phi(h)-\psi(h')& = f(h)-f(h') - D\|\nabla f(h)\|\\
    &  = f\circ g(1)-f\circ g(0) - D\|\nabla f(h)\|,
    \intertext{where $g(t)= th +(1-t)h'$. Then, by the fundamental theorem of calculus, we have }
    \phi(h)-\psi(h') & = \int_{0}^{1} (f\circ g)'(t) dt  - D\|\nabla f(h)\| \\
    & = \int_{0}^{1} \left\langle \nabla f\left( th+(1-t)h' \right), h-h' \right\rangle dt - D\|\nabla f(h)\|.
    \intertext{We now control the last term using that $\|h-h'\|\le D$ and Cauchy-Schwarz inequality:}
    \phi(h)-\psi(h')& \le \int_{0}^{1} \left\langle \nabla f\left( th+(1-t)h' \right), h-h' \right\rangle dt - \left\langle \nabla f(h), h-h' \right\rangle\\
    & =  \int_{0}^{1} \left\langle \nabla f\left( th+(1-t)h' \right) - \nabla f(h), h-h' \right\rangle dt.
    \intertext{Then by the $(\star)$ assumption:}
    \phi(h)-\psi(h')
    & \le \|h-h'\|\int_{0}^{1} (1-t) dt\left\| h-h'\right\| dt\\
    &= \frac{\|h-h'\|^2}{2}.
  \end{align*}
  We then conclude by applying \Cref{eq:kantorovich} to the pair $(2\phi,2\psi)$. The general case with $G\neq 1$ follows immediately by considering the pair $(\frac{2}{G}\phi,\frac{2}{G}\psi)$.
\end{proof}

\subsection{Proof of \Cref{cor:kl-change}}
\label{sec:proof-kl-change}

\corklchange*
\begin{proof}
  We fix $R>0$, and we start from \Cref{th:2-wass} with predictor space $\H_0=\Bcal(\zerobf,R)$, where $f$ is gradient-Lipschitz on this ball and the prior and the posterior are respectively $\Pcal_R\#\Q$ and $\Pcal_R\#\P$. We have
  \begin{align*}
  \EE_{h\sim \Q}\left[ f\left(\Pcal_R(h)\right) \right] \le \frac{G}{2}W_2^2\left(\Pcal_R\#\Q,\Pcal_R\#\P \right) + \EE_{h\sim \P}\left[ f\left(\Pcal_R(h)\right) \right] + 2R\EE_{h\sim \Q} \left[ \left\|\nabla_h f \left(\Pcal_R(h)  \right) \right\| \right].
  \end{align*}
  We first prove that $W_2^2\left(\Pcal_R\#\Q,\Pcal_R\#\P \right) \le W_2^2(\Q,\P)$. Let $\pi\in \Gamma(\Q,\P)$ be the optimal transport coupling from $\P$ to $\Q$, \ie 
  \begin{align*}
  W_2^2(\Q,\P)= \EE_{(X,Y)\sim \pi} \left[\|X-Y\|^2\right].
  \end{align*}
  Then, notice that if we denote by $\pi_1= (\Pcal_R,\Pcal_R)\#\pi$, then $\pi_1\in \Gamma\left( \Pcal_R\#\Q, \Pcal_R\#\P\right)$ and so:
  \begin{align*}
    W_2^2\left(\Pcal_R\#\Q,\Pcal_R\#\P \right) & \le \EE_{(X,Y)\sim \pi_1}\left[ \|X-Y\|^2 \right] \\
    & = \EE_{(X,Y)\sim \pi_1}\left[ \|\Pcal_R(X)-\Pcal_R(Y)\|^2 \right].
    \intertext{Using the fact that $\Pcal_R$ is $1$-Lipschitz gives:}
    W_2^2\left(\Pcal_R\#\Q,\Pcal_R\#\P \right) & \le \EE_{(X,Y)\sim \pi_1}\left[ \|X-Y\|^2 \right]\\
    & = W_2^2(\Q,\P).
  \end{align*}
  Next, we need to control $W_2^2(\Q,\P)$. 
  To do so, we use the fact that $\P$ is $\Lsob(c_{LS})$ to assert, through Otto-Villani's theorem \citep[Theorem 1]{otto2000generalization} that the following holds: $W_2^2(\Q,\P)\le \frac{c_{LS}(\P)}{2}\KL(\Q,\P)$. This concludes the proof. 
\end{proof}

\subsection{Proof of \Cref{th:wpb-grad}}
\label{sec:proof-wpb-grad}

\thwpbgrad*
\begin{proof}
    We start from \Cref{th:2-wass}, using the fact that $\Risk_\D(h)-\hat{\Risk}_{\S_m}(h)$ is $G$-gradient-Lipschitz for any $m\in\N^{*}$ to obtain:
    \begin{multline*}
        \EE_{h\sim \Q}[\Risk_\D(h)-\hat{\Risk}_{\S_m}(h)] \le \frac{G}{2}W_2^2(\Q,\P) + \EE_{h\sim \P}[\Risk_\D(h)-\hat{\Risk}_{\S_m}(h)]\\
        +D \EE_{h\sim \Q}\left( \left\| \nabla_h \Risk_\D(h) - \nabla_h \hat{\Risk}_{\S_m}(h) \right\| \right)
    \end{multline*}
    The only remaining term to control is $\EE_{h\sim \P}[\Risk_\D(\Q)-\hat{\Risk}_{\S_m}(\Q)]$. For this, we use the supermartingale concentration inequality of \citet[Corollary 17]{chugg2023unified} instantiated with the prior equals to the posterior, which shows that, for any $\lambda >0$, with probability at least $1-\delta$, we have
    \begin{align*}
    \EE_{h\sim \P}[\Risk_\D(h)-\hat{\Risk}_{\S_m}(h)] \le \frac{\log(1/\delta)}{\lambda} + \frac{\lambda}{2}\EE_{h\sim\P}\EE_{\z\sim\D}[\ell(h,\z)^2].
    \end{align*}
    The last term on the right-hand side is bounded by $\sigma^2$ by assumption. Taking $\lambda= \sqrt{\frac{2\log(1/\delta)}{\sigma^2}}$, we finally get $\EE_{h\sim\P}\EE_{\z\sim\D}[\ell(h,\z)^2] \le \sqrt{2\log(1/\delta)/m}$, which concludes the proof.
\end{proof}

\end{document}